%% file: main.tex
\documentclass[conf]{new-aiaa} 
\pdfoutput=1

\usepackage[utf8]{inputenc} 
\usepackage[T1]{fontenc}    

\usepackage[numbers]{natbib} 

\usepackage{graphicx}       
\usepackage{subcaption}     
\usepackage{rotating}       

\usepackage{amsmath, bm}    
\usepackage{siunitx}        
\usepackage{booktabs}       

\usepackage[version=4]{mhchem} 

\usepackage{longtable, tabularx} 
\usepackage{array}              

\usepackage{algorithm}      
\usepackage{algpseudocode}  

\usepackage[acronyms, shortcuts]{glossaries}

\usepackage{hyperref}       
\hypersetup{
    colorlinks=true,        
    linkcolor=black,        
    citecolor=black,        
    filecolor=blue,         
    urlcolor=blue           
}

\newcommand{\revised}[1]{{\leavevmode\color{black}#1}}

\input{definitions}

\title{Diffusion Policies for Generative Modeling of Spacecraft Trajectories}

\author{Julia Briden\footnote{Advanced Mission Design and GN\&C Engineer, Amentum, NASA Johnson Space Center, 2101 E NASA Pky, Houston, TX 77058 USA, and AIAA Member.}
}

\author{Breanna Johnson \footnote{Aerospace Engineer, Flight Mechanics and Trajectory Design Branch, EG5, NASA Johnson Space Center, 2101 E NASA Pky, Houston, TX, Senior Member AIAA.}}

\affil{2101 E NASA Pky, Houston, TX, 77058 USA}

\author{Richard Linares \footnote{Rockwell International Career Development Professor and Associate Professor, Department of Aeronautics and Astronautics, 125 Massachusetts Avenue. Senior Member AIAA.}}

\affil{Department of Aeronautics and Astronautics, Massachusetts Institute of Technology, 77 Massachusetts Avenue, Cambridge, Massachusetts, 02139 USA}

\author{Abhishek Cauligi \footnote{Robotics Technologist, Jet Propulsion Laboratory, California Institute of Technology, Pasadena, CA 91109, USA.}}

\affil{NASA Jet Propulsion Laboratory, California Institute of Technology, 4800 Oak Grove Dr, Pasadena, CA 91109 USA}

\begin{document}

\maketitle

\begin{abstract}
Machine learning has demonstrated remarkable promise for solving the trajectory generation problem and in paving the way for online use of trajectory optimization for resource-constrained spacecraft.
However, a key shortcoming in current machine learning-based methods for trajectory generation is that they require large datasets and even small changes to the original trajectory design requirements necessitate retraining new models to learn the parameter-to-solution mapping.
In this work, we leverage compositional diffusion modeling to efficiently adapt out-of-distribution data and problem variations in a few-shot framework for 6 degree-of-freedom (DoF) powered descent trajectory generation.
Unlike traditional deep learning methods that can only learn the underlying structure of one specific trajectory optimization problem, diffusion models are a powerful generative modeling framework that represents the solution as a probability density function (PDF) and this allows for the composition of PDFs encompassing a variety of trajectory design specifications and constraints.
We demonstrate the capability of compositional diffusion models for inference-time 6 DoF minimum-fuel landing site selection and composable constraint representations.
Using these samples as initial guesses for  6~DoF powered descent guidance enables dynamically feasible and computationally efficient trajectory generation.
\end{abstract}

\section{Nomenclature}

{\renewcommand\arraystretch{1.0}
\noindent\begin{longtable*}{@{}l @{\quad=\quad} l@{}}
$\alpha$ & negation parameter \\
$\alpha_t$ & $\Pi_{t=0}^T \beta_t$ \\
$\bar \alpha_t$ & $\Pi_{t=0}^T (1-\beta_t)$ \\
$\beta_t$ & forward process Gaussian noise variance \\
$\tilde\beta_t$ & $\sqrt{1-\beta_t}$ \\
$\Delta$ & deviation from previous iterates \\
$\delta$ & score function error \\
$\delta_{\max}$ & maximum gimbal angle \\
$\epsilon$ & standard normal noise \\
$\epsilon_\theta$ & score function \\
$\gamma_{gs}$ & glideslope angle \\
$\lambda_\text{penalty}$ & energy penalty \\
$\mu_\theta$ & backward process mean parameterized by $\theta$ \\
$\nu$ & virtual control \\
$\bm{\Omega}_{\omega_\mathcal{B} (t)}$ & quaternion transform \\
$\omega$ & penalty coefficients \\
$\bm{\omega}_\mathcal{B}$ & angular velocity \\
$\Sigma$ & covariance \\
$\sigma^i$ & time-scaling factor \\
$\sigma_t$ & $\sqrt{1 - \bar \alpha_t}$ \\
$\theta$ & learnable parameter \\
$\bar A$ & state dynamics matrix \\
$\mathbf{a}$ & action \\
$\bar B$ & control dynamics matrix \\
$\mathcal{B}$ & body frame \\
$b$ & batch size \\
$\bar C$ & future control dynamics matrix \\
$C_{\mathcal{I} \leftarrow \mathcal{B}}$
& transformation matrix \\
$c$ & cost distribution \\
$d$ & dimensionality of $\bm{x}_t$ \\
$\bm{e}$ & inertial frame unit vector $(e_1, e_2, e_3)$, defined at the landing site, with $e_1$ pointing in the opposite direction of $\bm{g}_\mathcal{I}$ \\
$E_\theta$ & energy function \\
$\bm{g}_\mathcal{I}$ & gravitational acceleration \\
$\mathbf{g}_t$ & state constraint at timestep $t$ \\
$H_\gamma$ & glideslope matrix \\
$h$ & conditioning function \\
$I$ & identity \\
$\mathcal{I}$ & inertial frame \\
$\mathcal{J}$ & cost function \\
$\mathbf{J}_\mathcal{B}$ & moment of inertia \\
$K$ & final time step \\
$m$ & mass \\
$m_\text{wet}$ & wet mass \\
$\mathcal{N}$ & normal distribution \\
$N$ & planning horizon \\
$p$ & backward process probability density function \\
$q$ & forward process probability density function \\
$\bm{q}_{\mathcal{B} \leftarrow \mathcal{I}}$ & quaternion \\
$R$ & risk level \\
$\bm{r}_\mathcal{I}$ & position \\
$\mathbf{s}$ & state \\
$t$ & diffusion timestep \\
$t_f$ & final time \\
$T$ & number of diffusion timesteps \\
$\mathbf{T}_\mathcal{B}$ & thrust \\
$T_{\max}$ & maximum thrust \\
$T_{\min}$ & minimum thrust \\
$\mathbf{u}$ & control inputs \\
$\bm{v}_\mathcal{I}$ & velocity \\
$\mathbf{x}_t$ & samples at diffusion timestep $t$ \\
$\bm{z}$ & states and control inputs \\
$\bar z$ & disturbance
\end{longtable*}}

\section{Introduction}
\lettrine{T}{rajectory} optimization has emerged as a powerful modeling paradigm to design spacecraft trajectories that are both dynamically feasible for the system and also satisfy the mission and science constraints~\cite{Betts1998,LiuLuEtAl2017}.
Despite its promise and the tremendous advances in nonlinear optimization solvers in recent years, trajectory optimization has primarily been constrained to offline usage due to the limited compute capabilities of radiation hardened flight computers~\cite{SoleWolfEtAl2023}.
However, with a flurry of proposed mission concepts that call for increasingly greater on-board autonomy~\cite{StarekAcikmeseEtAl2016}, bridging this gap in the state-of-practice is necessary to allow for scaling current trajectory design techniques for future missions.

Recently, researchers have turned to machine learning and data-driven techniques as a promising method for reducing the runtimes necessary for solving challenging constrained optimization problems~\cite{KotaryFiorettoEtAl2021,Amos2023}.
Such approaches entail learning what is known as the problem-to-solution mapping between the problem parameters that vary between repeated instances of solving the trajectory optimization problem to the full optimization solution and these works typically use a~\ac{dnn} to model this mapping~\cite{Sambharya2023EndToEnd,briden2024improving,CauligiChakrabartyEtAl2022}.
Given parameters of new instances of the trajectory optimization problem, this problem-to-solution mapping can be used online to yield candidate trajectories to warm start the nonlinear optimization solver and this warm start can enable significant solution speed ups.
One shortcoming of these aforementioned data-driven approaches is that they have limited scope of use and the learned problem-to-solution mapping only applies for one specific trajectory optimization formulation.
With a change to the mission design specifications that yields, \eg{}, a different optimization constraint, a new problem-to-solution mapping has to be learned offline and this necessitates generating a new dataset of solved trajectory optimization problems.
To this end, our work explores the use of compositional diffusion modeling to allow for generalizable learning of the problem-to-solution mapping and equip mission designers with the ability to interleave different learned models to satisfy a rich set of trajectory design specifications.

\begin{figure}
    \centering
    \includegraphics[width=0.7\textwidth]{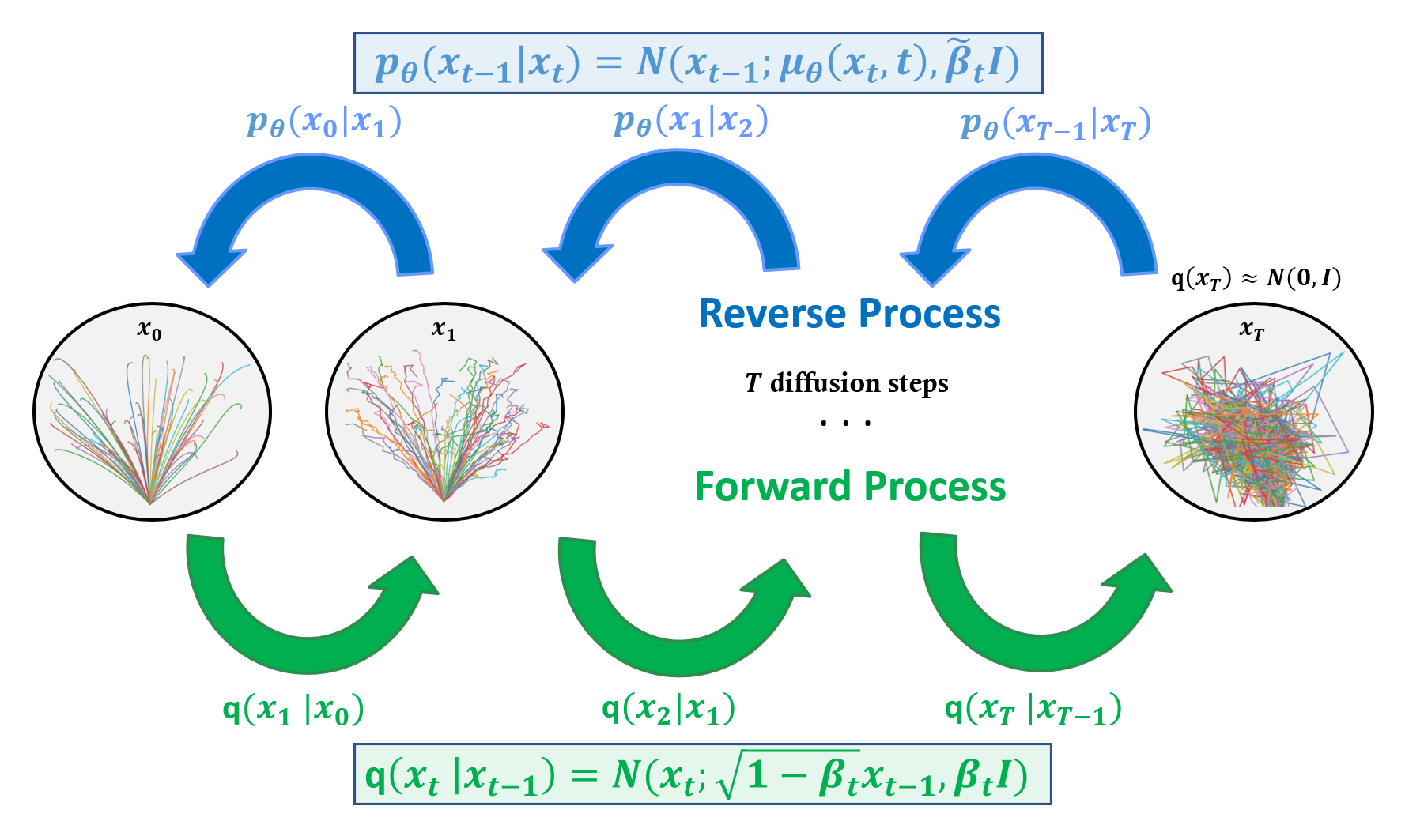}
    \caption{Forward and reverse processes for a generative diffusion model.
    }
    \label{fig:forwardreverseprocess}
\end{figure}

Compositional diffusion modeling enables training of a model to both sample and plan from.
By learning a sequential perturbation, $\mathbf{\epsilon}_\theta (\mathbf{x}_t, t)$ for sample $\mathbf{x}_t$ at diffusion timestep $t$, diffusion models recover a representative distribution of trajectory samples from a standard normal distribution (Figure~\ref{fig:forwardreverseprocess}).
Unlike directly training a~\ac{dnn} on trajectory data, learning this perturbation over the diffusion time steps enables accurate estimation in regions of low data density \cite{song2019generative, song2020improved}.
Furthermore, exploiting the energy-based formulation of diffusion models allows diffusion models to be directly composed during inference time, generalizing beyond training data for multiple tasks~\cite{JannerDuEtAl2022, du2024compositional, briden2024compositional}.
Using trajectory-level diffusion probabilistic models, all timesteps of a plan can be predicted simultaneously. The sampling procedure, the reverse process in Figure~\ref{fig:forwardreverseprocess}, can then recover trajectories that satisfy imposed constraints by conditioning on state and control constraints~\cite{JannerDuEtAl2022}. Since generated trajectories are over long horizons, diffusion probabilistic models do not have the compounding rollout errors of single-step dynamics models \cite{JannerDuEtAl2022}.
By iteratively improving local consistency, models can generalize to new settings by combining in-distribution subsequences~\cite{du2024compositional, briden2024compositional}.

\subsection{Related Work}
Trajectory optimization is a powerful tool to generate dynamically feasible trajectories for autonomous spacecraft planning, but existing radiation-hardened computers used in aerospace applications lack the computational resources to be able to run these at necessary rates for online use.
To reduce the runtime requirements for such algorithms, researchers have turned towards machine learning-based techniques to accelerate solution times for trajectory optimization problems of interest.
A common approach has involved using a neural network to learn the mapping between the problem parameters that vary between repeated instances of solving the trajectory optimization problem to the optimal solution attained after running trajectory optimization.
Such warm starting techniques have been applied to both~\acp{scp} and~\acp{sqp}-based approaches and have demonstrated solution speeds up of up to two orders of magnitude~\cite{Chen2022LargeScale, CauligiChakrabartyEtAl2022, Sambharya2023EndToEnd, briden2024improving}.

Recently, trajectory generation has turned to generative models, including diffusion models, to generate feasible trajectories and control inputs from a distribution~\cite{JannerDuEtAl2022, Ajay2022IsConditional, Chi2023DiffusionPolicy, Carvalho2023MotionPlanning, Sridhar2023Nomad}.
While the compositional feature of generative models enables the integration of cost functions and constraints into these models without additional training, current work explores the integration of safety constraints into generative models while balancing solution accuracy, efficiency, and verification methods~\cite{Li2023Amortized, Chang2023Denoising, Yang2023Compositional, Xiao2023SafeDiffuser, Power2023Sampling, Botteghi2023Trajectory}.
A less algorithmically intrusive method of ensuring constraint satisfaction includes sampling from generative models to warm start the numerical solver for~\acp{nlp}~\cite{Li2024Efficient}.
By applying a guided sampling strategy for a trained diffusion model, the solve time for a tabletop manipulation task was over 22 times faster than the strategy produced by a uniform sampling method, and using only an unconstrained diffusion sampled initial guess solved the cislunar transfer problem over 11 times faster than the uniform method~\cite{Li2024Efficient}.
C-TrajDiffuser utilized product and negation composition to generate powered descent trajectory samples that conform to the state-triggered constraints and atmospheric drag in the dynamics~\cite{briden2024compositional}.
When parameterizing the control for the dynamical system using a neural network and training it as a stochastic optimization algorithm on a loss function,~\ac{ido} techniques are used \cite{domingo-enrich2024stochastic}.
Previous work by Domingo-Enrich et al. establishes~\ac{socm} as an~\ac{ido} technique that fits a random matching vector field to perform path-wise reparameterization.

\begin{figure}
    \centering
    \includegraphics[width=0.9\textwidth]{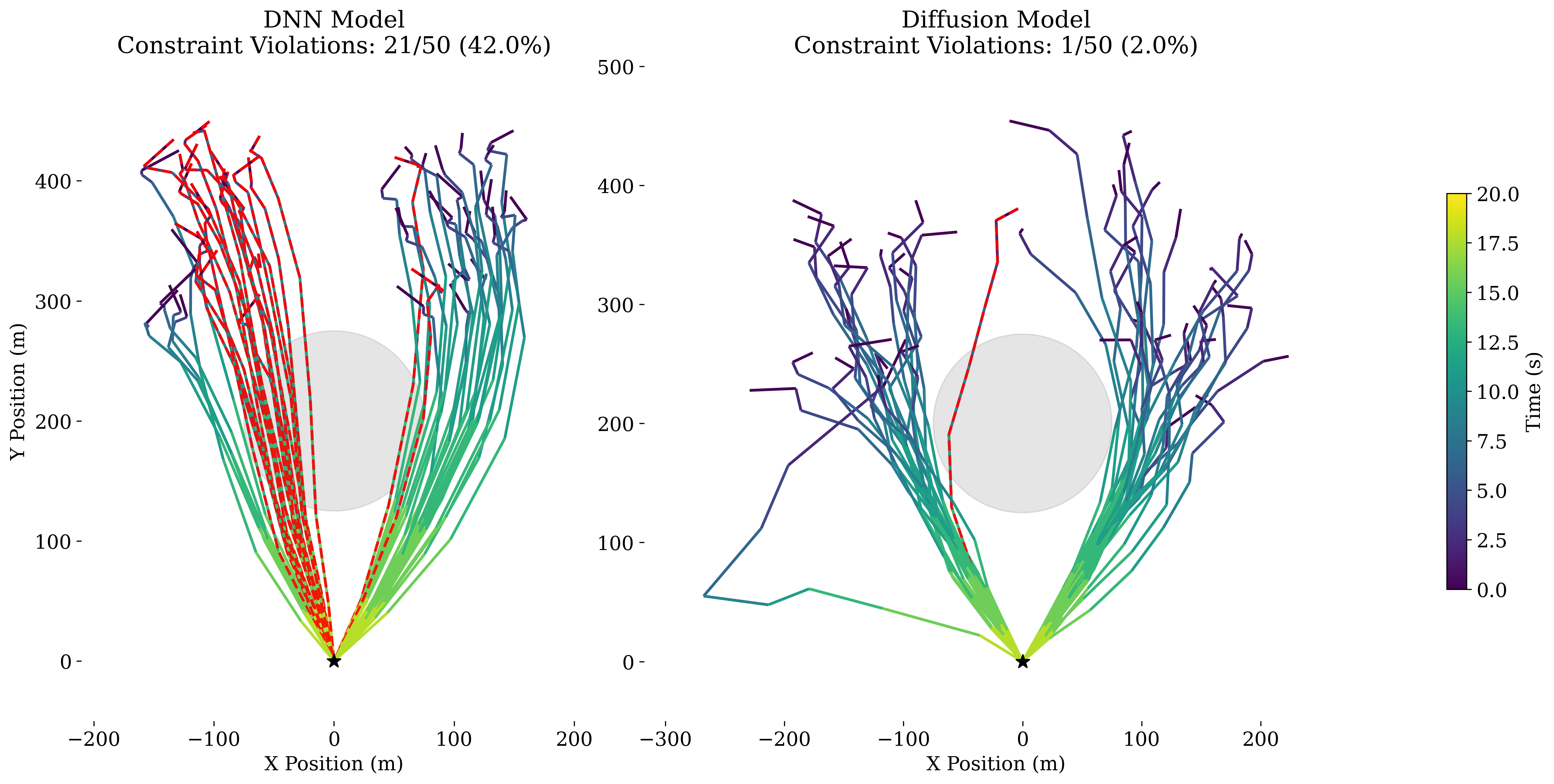}
    \caption{Sampled trajectories for 2D  powered descent guidance around an obstacle, the gray circle.
    As this example illustrates, diffusion models are a powerful policy representation that can efficiently learn the inherently multi-model structure of the underlying task and thereby significantly reduce constraint violation compared to a ``vanilla''~\ac{dnn} approach.
    }
    \label{fig:trajcomparison}
\end{figure}

A crucial difference between diffusion-learned trajectory generation and trajectories generated by~\acp{dnn} is the handling of multimodal distributions.
In particular, the stochastic sampling and initialization processes for~\acp{ddpm} via Gaussian perturbations promote convergence to multi-modal action basins~\cite{Chi2023DiffusionPolicy}.
In contrast,~\acp{dnn} fit continuous functions to the dataset via explicit modeling, leading to interpolated values when out-of-distribution inputs are utilized~\cite{FlorenceLynchEtAl2022}.
We illustrate this in a pedagogical example shown in Figure~\ref{fig:trajcomparison}, where a spacecraft must choose between two homotopy classes corresponding to moving either to the left or right of the obstacle.
In such a case of combinatorial decision making, conventional learned trajectory optimization techniques struggle~\cite{SrinivasanChakrabartyEtAl2020}, with the feedforward~\ac{dnn} violating the collision avoidance constraint in 42\% of cases as compared to only 2\% for the diffusion model trained on the same dataset.
While previous work on diffusion models and guided sampling for trajectory generation demonstrates the efficacy and accuracy of the method for these problem types, only a few applications have leveraged the compositional nature of diffusion models to generalize to alternate constraint formulations for warm-starting constrained optimization problems.

\subsection{Contributions}

In this work, we generate efficient and generalizable trajectories for the 6~\ac{dof} powered descent guidance problem using a trained diffusion model.
By composing the learned trajectory diffusion model with the distribution of low-risk landing sites, generated samples efficiently solve the multi-landing site selection problem for trajectory generation.
The dynamics diffusion model is trained to effectively model the probability density distribution of 6~\ac{dof} powered descent guidance with an unconstrained landing location.
Constraint satisfaction can be achieved during inference time by composing the trajectory diffusion model together with a differentiable glideslope energy-based diffusion model.
State and control conditioning enables a flexible alternative to input trajectory constraints.
To the best of the authors' knowledge, this work is the first use of generative composition and multi-modality for high-dimensional multi-landing site trajectory generation problems.

\section{Background}

Given a discrete-time dynamics system, with each state defined as $\mathbf{s}_{t+1} = f(\mathbf{s}_t, \mathbf{a}_t)$. Direct optimization problems seek to minimize a cost $\mathcal{J}$ function over costs $c(\mathbf{s}_t, \mathbf{a}_t)$, subject to constraints on the states and actions:

\begin{equation}
    \mathbf{a}^*_{0:N} = \arg \min_{\mathbf{a}_{0:N}} \mathcal{J}(\mathbf{s}_0, \mathbf{a}_{0:N}) = \arg \min_{\mathbf{a}_{0:N}} \sum_{t=0}^N c(\mathbf{s}_t, \mathbf{a}_t)
    \label{eq: constrained opt}
\end{equation}

\[
\text{subject to} \quad \mathbf{s}_t \in \mathcal{S}, \quad \mathbf{a}_t \in \mathcal{A}, \quad \forall t = 0, 1, \ldots, N.
\]

where $\mathcal{S}$ and $\mathcal{A}$ denote the feasible sets of states and actions, respectively. In this work, we define a trajectory as a two-dimensional array:

\begin{equation}
    \mathbf{x} = \begin{bmatrix}
    \mathbf{s}_0 & \mathbf{s}_1 & \cdots & \mathbf{s}_N \\
    \mathbf{a}_0 & \mathbf{a}_1 & \cdots & \mathbf{a}_N
    \end{bmatrix}.
    \label{eq: trajectory}
\end{equation}

While directly learning these trajectories or the active constraints at the optimal solution enables accurate approximations for the solution to Equation~\eqref{eq: constrained opt}, a simple change in the cost function or one constraint may yield an entirely different solution mapping, requiring resampling and retraining of the learned mapping.
In contrast, diffusion probabilistic models allow for unconditioned dynamics models to be learned via an iterative denoising process, which can be conditioned on the desired cost functions and constraints during inference time \cite{du2024compositional, Chi2023DiffusionPolicy, briden2024compositional}.

\subsection{Diffusion Generative Probabilistic Models}

Diffusion probabilistic models generate data through an iterative denoising procedure $p_\theta (\mathbf{x}_{t-1} | \mathbf{x}_{t})$, reversing the forward diffusion process described by $q(\mathbf{x}_t | \mathbf{x}_{t-1})$ (Figure~\ref{fig:forwardreverseprocess}).
The data distribution for the noiseless data trajectory is given by Equation~\eqref{eq: data dist}:

\begin{equation}
    p_\theta (\mathbf{x}_0) = \int p(
    \mathbf{x}_N) \Pi_{t=1}^T p_\theta (\mathbf{x}_{t-1} | \mathbf{x}_t) d \mathbf{x}_{1:N}
    \label{eq: data dist}
\end{equation}

where $p(\mathbf{x}_T) = \mathcal{N} (0, \mathbf{I})$ is a standard Gaussian prior and $\mathbf{x}_0$ is the noiseless data.

Minimization of a variational bound on the negative log-likelihood of the reverse process gives the optimal parameters: $\theta^* = \arg \min_\theta - \mathbb{E}[\log p_\theta (\mathbf{x}_0)]$, where the reverse process is parameterized as a Gaussian with timestep-dependent covariances:

\begin{equation}
    p_\theta (\mathbf{x}_{t-1} | \mathbf{x}_t) = \mathcal{N} (\mathbf{x}_{t-1} | \mu_\theta (\mathbf{x}_t, t), \Sigma_t).
\end{equation}

\subsubsection{Training}

To train the score function $\epsilon_\theta$, samples are first computed using the \textit{forward process} at every diffusion timestep $t$, with a user specified Gaussian noise variance $0 < \beta_t \leq 1$:

\begin{equation}
    {\bm{x}}_t ({\bm{x}}_0, {\bm{\epsilon}}) = \sqrt{1-\sigma_t^2} {\bm{x}}_0 + \sigma_t {\bm{\epsilon}},
\end{equation}

where $\bm{\epsilon} \sim {\mathcal{N}} (0, I)$, $\sigma_t = \sqrt{1 - \bar \alpha_t}$, and $\bar \alpha_t = \Pi_{t=0}^T (1-\beta_t)$.
Next, the variational bound on the marginal likelihood for the reverse process is maximized by training the deep neural network $\bm{\epsilon}_\theta$ with the loss function in Equation~\eqref{eqn:loss_diffusion}~\cite{Botteghi2023Trajectory}.

\begin{equation}
    {\mathcal{L}} ({\bm{\theta}}) = \sum_{t=1}^T {\mathbb{E}}_{q({\bm{x}}_0) {\mathcal{N}} ({\bm{\epsilon}}; 0, I) } \left[ \| {\bm{\epsilon}} - {\bm{\epsilon}}_{\bm{\theta}} ({\bm{x}}_t ({\bm{x}}_0, {\bm{\epsilon}}), t) \|^2 \right].
    \label{eqn:loss_diffusion}
\end{equation}

We refer the reader to~\cite{briden2024compositional} for a short derivation of Equation~\eqref{eqn:loss_diffusion} from the variational bound on the marginal likelihood for the reverse process.

\subsubsection{Sampling}

Once the score function $\bm{\epsilon}_\theta$ is trained, we can now use the \textit{reverse process} to sample from our trajectory distribution.
First, we draw one or more samples from the standard normal distribution ${\bm{x}}_T \sim p({\bm{x}}_T) \approx \mathcal{N}$$( 0, I )$.
We then sample ${\bm{x}}_{t-1}$ for $T$ diffusion steps using Equation~\eqref{eqn:diffusion_sample} until $t=0$, the sample from the original trajectory distribution, is achieved.
\begin{equation}
    {\bm{x}}_{t-1} = {\bm{\mu}}_{\bm{\theta}} ({\bm{x}}_{t}, t) + \sqrt{\tilde \beta_t} {\bm{\epsilon}},
    \label{eqn:diffusion_sample}
\end{equation}
where ${\bm{\epsilon}} \sim \mathcal{N}$$( 0, I )$ and ${\bm{\mu}}_{\bm{\theta}} = \frac{1}{\sqrt{\alpha_t}} \left({\bm{x}}_t - \frac{\beta_t}{\sqrt{1 - \bar \alpha_t}} {\bm{\epsilon}}_{\bm{\theta}} ({\bm{x}}_t, t)\right)$.
This process is known as reverse diffusion sampling and is equivalent to annealed~\ac{mcmc}, or~\ac{ula}~\ac{mcmc} sampling with an implicit gradient represented by the score function 
$\bm{\epsilon}_\theta$ \cite{du2019implicit, roberts1996exponential}.
This implicit gradient representation will be further explained and utilized in the following section.

\subsection{Compositional Diffusion Models}

If we consider the diffusion model as parameterizing a~\ac{pdf} $p_\theta$ for a distribution that serves as a trajectory generator, assuming model independence, we can compose learned diffusion models.
Given the learned parameter $\theta$, the data distribution can then be perturbed to achieve flexible conditioning using information about prior evidence and required outcomes, given by $h(\mathbf{x})$:

\begin{equation}
    \tilde{p}_\theta(\mathbf{x}) \propto p_\theta (\mathbf{x}) h(\mathbf{x}).
    \label{eq: product}
\end{equation}

Leveraging distributions with independent structures between variables substantially reduces the data required to learn complex distributions.
Further, the learned joint distribution can be generalized to unseen combinations of variables under only local variable in-distribution requirements~\cite{du2024compositional}.

While Equation~\eqref{eq: product} demonstrates a product compositional distribution, mixtures 

\begin{equation}
    \tilde{p}_\theta(\mathbf{x}) \propto p_\theta (\mathbf{x}) + h(\mathbf{x}),
\end{equation}

and negation 

\begin{equation}
    \tilde{p}_\theta(\mathbf{x}) \propto \frac{p_\theta (\mathbf{x})}{h(\mathbf{x})^\alpha},
\end{equation}

provide additional flexibility for composition options~\cite{du2024compositional}.
The resultant~\acp{pdf} are proportional, rather than equal, due to the absence of the normalization constant which is assumed to be one in most cases due to computational intractability for actually computing it~\cite{du2024compositional}.

Since composition requires an explicit representation of the probability density function, differentiable~\acp{ebm}~\cite{du2023reduce} are utilized to model the negative log-likelihood for a diffusion model, where the sequence of learned energy functions $E_\theta$ are proportional to the probability density function:

\begin{equation}
    p_{\bm{\theta}}^t(\bm{x}) \propto e^{-E_{\bm{\theta}}(\bm{x}, t)}.
    \label{eq: EBM}
\end{equation}

We can then obtain the score function to use in the reverse process by differentiating the energy function $E_\theta$ at the desired diffusion timestep $t$:

\begin{equation}
    {\bm{\epsilon}}_{\bm{\theta}}(\bm{x}_t, t) = \frac{\nabla E_{\bm{\theta}}(\bm{x}_t, t)}{\sqrt{1 - \bar \alpha_t}}.
    \label{eq: EBM score}
\end{equation}

We note that the reverse diffusion process, defined by~\ac{ula}~\ac
{mcmc}, does not sample from the exact composed distribution and a trade-off between sample accuracy  and computational efficiency exists~\cite{du2023reduce}.
Specifically, the upper bound on score function error for product composition $\delta_{\text{prod}} ({\bm{x}},t) = |{\bm{\epsilon}}_{{\bm{\theta}} \; \text{prod}} ({\bm{x}},t) - \tilde {\bm{\epsilon}}_{{\bm{\theta}} \; \text{prod}} ({\bm{x}},t)|$, is defined as follows:

\begin{align}
    \delta_{\text{prod}} ({\bm{x}}) 
    &\leq \sum_{t=1}^T \frac{1}{1-\bar \alpha_t} ((N-1) \sqrt{\bar \alpha_t {\text{Tr}} ({\text{Var}} ({\bm{x}}_0)) + (1-\bar{\alpha}_t) d}  \\
    &+ N \sqrt{\bar \alpha_t} {\sqrt{ 2 {\text{Tr}} ( {{\text{Var}}_{q_0}} ({\bm{x}}_0)})} ),
    \label{eq: bound}
\end{align}

where $d$ is the dimensionality of ${\bm{x}}_t$ (we refer the reader to~\cite{briden2024compositional} for a derivation of this bound.)
Including full~\ac{mcmc} iterations in the backward process for composed models improves sample accuracy~\cite{du2023reduce}, often with a cost of increased computational complexity.
Since all samples will be used as initial guesses to an optimizer, only reverse diffusion was used for this work.
Future work will explore the performance improvement and additional computational costs of using~\ac{mcmc} for compositional diffusion for trajectory generation.

An applicable example of energy-based diffusion for trajectory optimization is the composition of a probability density trained on the system dynamics $p_{\text{traj}}(\mathbf{\tau})$ and a probability density which specifies the start $\mathbf{s}_0$ and goal $\mathbf{s}_T$ states, or boundary conditions, $p_{\text{bc}}(\mathbf{\tau}, \mathbf{s}_0, \mathbf{s}_T)$.
Here, the planning distribution to sample from becomes $\tilde{p}_\theta(\mathbf{\tau}) \propto p_{\text{traj}}(\mathbf{\tau}) p_{\text{bc}}(\mathbf{\tau}, \mathbf{s}_0, \mathbf{s}_T)$.
Since $p_{\text{traj}}(\mathbf{\tau})$ denotes a distribution defining only the dynamics, it may be reused and conditioned for alternative cost functions and constraints without retraining.

\begin{figure}
    \centering
    \includegraphics[width=0.9\textwidth]{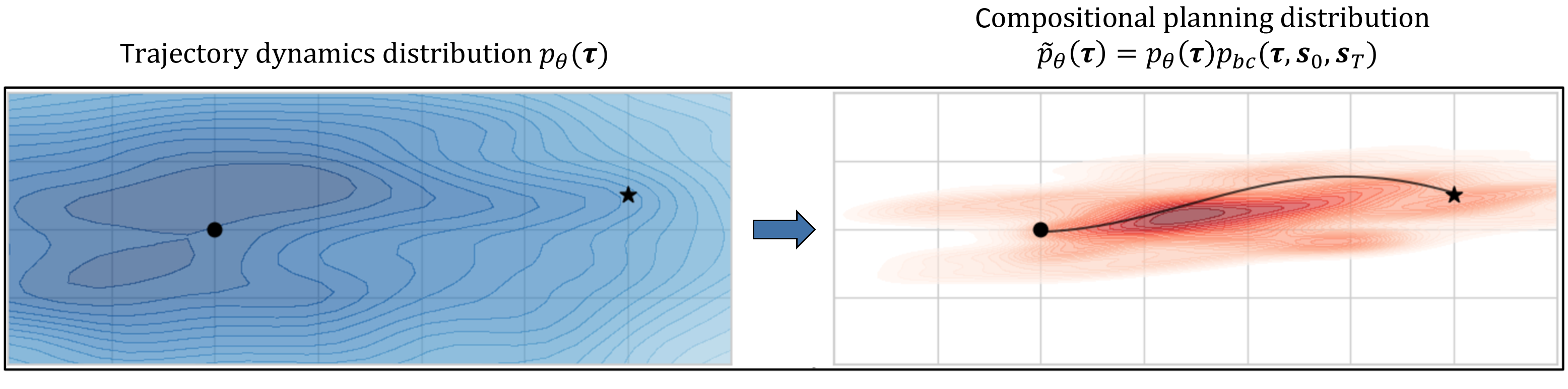}
    \caption{By composing the trajectory dynamics distribution, shown in blue, with the Dirac delta perturbation function for satisfaction of the start and goal constraints, the compositional planning distribution, shown in red, allows for constraint-satisfying sampling without retraining.}
    \label{fig:inpainting}
\end{figure}

In the context of trajectory planning problems, \textit{inpainting}, where state and action constraints act analogously to pixels in an image, produces feasible trajectories that satisfy a set of constraints (Figure~\ref{fig:inpainting}). 
The Dirac delta can then define the perturbation function for observed values and constant everywhere else.
Equation~\eqref{eq: state constraint} shows the state constraint implementation:

\begin{equation}
    h(\mathbf{\tau}) = \delta_{\mathbf{g}_t} (\mathbf{s}_0, \mathbf{a}_0, \dots, \mathbf{s}_T, \mathbf{a}_T) = \begin{cases}
    +\infty & \text{if } \mathbf{g}_t = \mathbf{s}_t \\
    0 & \text{otherwise}
    \end{cases}
    \label{eq: state constraint}
\end{equation}

where $\mathbf{g}_t$ is a state constraint at timestep $t$.
The same formulation holds for control constraints, where $\mathbf{a}_t$ instead of $\mathbf{s}_t$ determines constraint enforcement.
Figure~\ref{fig:inpainting} shows an example of inpainting for model composition.
On the left, the blue contours denote the distribution of sampled trajectories for simple 2D double integrator dynamics when composed with two state constraint distributions (Eq.~\eqref{eq: state constraint}), one on the start and one on the goal.
The resulting distribution in red illustrates the now tightly clustered distribution of product-composed constraint-satisfying trajectories.

\section{Methods}

In this work, we develop a generalized compositional diffusion model for powered descent guidance and evaluate its performance on out-of-distribution problem types by composing new models with simple and easy-to-train examples to generate more complex cost and constraint-satisfying models.
Further, the strengths of diffusion-based generative modeling will be shown for multi-modal problem scenarios, including the multi-landing site selection problem~\cite{hayner2023halo}.

The generalized compositional diffusion model is defined in Equation~\eqref{eq: general}:

\begin{equation}
    p_\theta (\mathbf{x}) \propto \Pi_i p_\theta^i (\mathbf{x}|\text{env}) \Pi_j c^j \Pi_k g^k,
    \label{eq: general}
\end{equation}

where $p_\theta (\mathbf{x} | \text{env})$ are independent diffusion models that depend on a specific environmental setup and/or a set of problem constraints, $g$ are the distributions for constraints that can be defined over the trajectory $\mathbf{x}$, and $c$ are the set of cost distributions over the trajectory.
Under an assumption of independence, these distributions can be composed to generate novel trajectories, distributed as $p_\theta (\mathbf{x})$, during inference time.

\subsection{Powered Descent Diffusion Model}

The 6-~\ac{dof} free-final-time powered descent guidance problem formulation used in this work assumes that speeds are sufficiently low such that planetary rotation and changes in the planet's gravitational field are negligible. 
The spacecraft is assumed to be a rigid body with a constant center-of-mass and inertia and a fixed center-of-pressure.
The propulsion is modeled as a single rocket engine that can be gimbaled symmetrically about two axes bounded by a maximum gimbal angle $\delta_\text{max}$.
The engine is assumed to be throttleable between $T_\text{min}$ and $T_\text{max}$, remaining on until the terminal boundary conditions are met.
The minimum-time 6~\ac{dof} powered descent guidance problem is formulated as follows:

\textbf{Cost Function:}
\begin{equation}
    \min_{t_f, \bm{T}_{{\mathcal{B}}(t)}} t_f
    \label{eq:cost}
\end{equation}
\textbf{Boundary Conditions:}
\begin{subequations}
\label{eq:boundary_conditions1}
\begin{align}
    & m(0) = m_{\text{wet}}\\
    & {\bm{r}}_{\mathcal{I}}(0) = r_{{\mathcal{I}}, z , i}  \\
    & {\bm{v}}_{\mathcal{I}}(0) = v_{{\mathcal{I}}, i}  \\
    & {\bm{\omega}}_{\mathcal{B}}(t_{0}) = {\bm{\omega}}_{{\mathcal{B}},i}\\
    & {\bm{r}}_{\mathcal{I}}(t_f) = {\bm{0}}\\
    & {\bm{v}}_{\mathcal{I}}(t_f) = {\bm{v}}_{{\mathcal{I}},f} \\
    & {\bm{q}}_{{\mathcal{B}} \leftarrow {\mathcal{I}}}(t_f) = {\bm{q}}_{{\mathcal{B}} \leftarrow {\mathcal{I}},  f}\\
    & {\bm{\omega}}_B(t_f) = 0 \\
    & {\bm{e_2}} \cdot {\bm{T}}_{\mathcal{B}} (t_f) = {\bm{e_3}} \cdot {\bm{T}}_{\mathcal{B}} (t_f) = 0
\end{align}
\end{subequations}
\textbf{Dynamics:}
\begin{subequations}
\label{eq:dynamics2}
\begin{align}
    & \dot{m}(t) = -\alpha_{\dot{m}} \|{\bm{T}}_{\mathcal{B}}(t) \|_2 \\
    & \dot{\bm{r}}_{\mathcal{I}}(t) = {\bm{v}}_{\mathcal{I}}(t)\\
    & \dot{\bm{v}}_{\mathcal{I}}(t) = \frac{1}{m(t)} {\bm{C}}_{{\mathcal{I}} \leftarrow {\mathcal{B}}}(t) {\bm{T}}_{\mathcal{B}}(t) + {\bm{g}}_{\mathcal{I}} \\
    & \dot{\bm{q}}_{{\mathcal{B}} \leftarrow {\mathcal{I}}}(t) = \frac{1}{2} \Omega_{{\bm{\omega}}_{\mathcal{B}}(t)} \bm{q}_{{\mathcal{B}} \leftarrow {\mathcal{I}}}(t) \\
    & {\bm{J}}_{\mathcal{B}} \dot{{\bm{\omega}}}_{\mathcal{B}}(t) = {\bm{r}}_{{\text{T}},{\mathcal{B}}} \times {\bm{T}}_{\mathcal{B}}(t) - {\bm{\omega}}_{\mathcal{B}}(t) \times {\bm{J}}_{\mathcal{B}} {\bm{\omega}}_{\mathcal{B}}(t)
\end{align}
\end{subequations}
\textbf{State Constraints:}
\begin{subequations}
\label{eq:state_constraints1}
\begin{align}
    & m_{\text{dry}} \leq m(t)
\end{align}
\end{subequations}
\textbf{Control Constraints:}
\begin{subequations}
\label{eq:control_constraints1}
\begin{align}
    & 0 < T_{\text{min}} \leq \|{\bm{T}}_{\mathcal{B}}(t)\|_2 \leq T_{\text{max}} \\
    & \cos \delta_{\text{max}} \|{\bm{T}}_{\mathcal{B}}(t)\|_2 \leq {\bm{e}}_1 \cdot {\bm{T}}_{\mathcal{B}}(t),
\end{align}
\end{subequations}

where the objective given in Equation~\eqref{eq:cost} minimizes the final time \( t_f \) while controlling the thrust vector \( \bm{T}_{\mathcal{B}}(t) \), which is expressed in the body frame \( \mathcal{B} \).
The boundary conditions in Eq.~\eqref{eq:boundary_conditions1} specify the initial and final states, where \( m(0) = m_{\text{wet}} \) represents the initial wet mass of the vehicle, \( {\bm{r}}_{\mathcal{I}}(0) = r_{{\mathcal{I}}, z, i} \) and \( {\bm{v}}_{\mathcal{I}}(0) = v_{{\mathcal{I}}, i} \) are the initial position and velocity in the inertial frame \( \mathcal{I} \), and \( \bm{\omega}_{\mathcal{B}}(t_0) = \bm{\omega}_{{\mathcal{B}}, i} \) denotes the initial angular velocity in the body frame.
At the final time, the vehicle reaches the desired final position \( \bm{r}_{\mathcal{I}}(t_f) = \bm{0} \), final velocity \( \bm{v}_{\mathcal{I}}(t_f) = \bm{v}_{{\mathcal{I}}, f} \), and orientation described by the quaternion \( \bm{q}_{\mathcal{B} \leftarrow \mathcal{I}}(t_f) = \bm{q}_{{\mathcal{B}} \leftarrow {\mathcal{I}}, f} \), with the angular velocity \( \bm{\omega}_{\mathcal{B}}(t_f) = 0 \).
To enable multi-landing site selection, only the z component of the final state is constrained to be zero, while the x and y components are sampled.
The thrust direction constraints ensure that the thrust components along certain axes are zero at the final time. 
The dynamics in Equation~\eqref{eq:dynamics2} are governed by the mass depletion rate \( \dot{m}(t) \), position evolution \( \dot{\bm{r}}_{\mathcal{I}}(t) \), velocity dynamics \( \dot{\bm{v}}_{\mathcal{I}}(t) \), quaternion kinematics \( \dot{\bm{q}}_{{\mathcal{B}} \leftarrow {\mathcal{I}}}(t) \), and angular velocity evolution governed by the moment of inertia \( \bm{J}_{\mathcal{B}} \) and the applied thrust.
Equations~\eqref{eq:state_constraints1} and~\eqref{eq:control_constraints1} ensure that the mass is bounded, a maximum tilt angle is not exceeded, and thrust is bounded.
For an extended discussion and derivation of the 6~\ac{dof} powered descent guidance problem and~\ac{scvx}, we refer the reader to~\cite{SzmukAcikmese2018, SzmukReynoldsEtAl2020}.

To solve the non-convex minimization problem posed in Equations~\eqref{eq:cost}-\eqref{eq:control_constraints1}, a series of convex~\ac{socp} problems can be solved until convergence in a process known as~\ac{scvx}.
If the non-convex problem admits a feasible solution, the converged solution will be constraint-satisfying.
The~\ac{socp} sub-problem that is iteratively solved during~\ac{scvx} is defined according to

\textbf{Cost Function:}
\begin{equation}
    \min_{\sigma^i, {\bm{u}}_k^i} \sigma^i + {\bm{\omega}}_\nu \|{\bm{\bar \nu}}^i\|_1 + {\bm{\omega}}_{{\bm{\Delta}}}^i \| {\bm{\bar {\bm{\Delta}}}}^i \|_2 + {\bm{\omega}}_{{\bm{\Delta}}_{\sigma}} \| {\bm{\Delta}_{\sigma}} \|_1
    \label{eq:cost_function}
\end{equation}
\textbf{Boundary Conditions:}
\begin{subequations}
\label{eq:boundary_conditions2}
\begin{align}
    & m_0^i = m_{\text{wet}} \\
    & {\bm{r}}_{{\mathcal{I}}, 0}^i = {{\bm{r}}}_{{\mathcal{I}}, i} \\
    & {\bm{v}}_{{\mathcal{I}}, 0}^i = {{\bm{v}}}_{{\mathcal{I}}, i} \\
    & {\bm{\omega}}_{{\mathcal{B}}, 0}^i = {\bm{\omega}}_{{\mathcal{B}}, i} \\
    & {\bm{r}}_{{\mathcal{I}}, K}^i = 0 \\
    & {\bm{v}}_{{\mathcal{I}}, K}^i = {\bm{v}}_{{\mathcal{I}}, f} \\
    & {\bm{q}}_{{\mathcal{B}} \leftarrow {\mathcal{I}}, K}^i = {\bm{q}}_{{\mathcal{B}} \leftarrow {\mathcal{I}}, f} \\
    & {\bm{\omega}}_{{\mathcal{B}}, K}^i = 0 \\
    & {\bm{e}}_2 \cdot {\bm{u}}_{{\mathcal{B}}, K}^i = {\bm{e}}_3 \cdot {\bm{u}}_{{\mathcal{B}}, K}^i = 0
\end{align}
\end{subequations}
\textbf{Dynamics:}
\begin{equation}
    {\bm{x}}_{k+1}^i = {\bar {\bm{A}}}_k^i {\bm{x}}_k^i + {{\bar {\bm{B}}}}_k^i {\bm{u}}_k^i + {{\bar {\bm{C}}}}_k^i {\bm{u}}_{k+1}^i + \bar \Sigma_k^i \sigma^i + {\bar {\bm{z}}}_k^i + \nu_k^i
    \label{eq:dynamics1}
\end{equation}
\textbf{State Constraints:}
\begin{subequations}
\label{eq:state_constraints2}
\begin{align}
    & m_{\text{dry}} \leq m_k^i
\end{align}
\end{subequations}
\textbf{Control Constraints:}
\begin{subequations}
\label{eq:control_constraints2}
\begin{align}
    & T_{\text{min}} \leq B_{\bm{g}}(\tau_k) {\bm{u}}_k^i \\
    & \| {\bm{u}}_k^i \|_2 \leq T_{\text{max}} \\
    & \cos \delta_{\text{max}} \| {\bm{u}}_k^i \|_2 \leq {\bm{e}}_1 \cdot {\bm{u}}_k^i
\end{align}
\end{subequations}
\textbf{Trust Regions:}
\begin{subequations}
\label{eq:trust_regions}
\begin{align}
    & \delta {\bm{x}}_k^i \cdot \delta {\bm{x}}_k^i + \delta {\bm{u}}_k^i \cdot \delta {\bm{u}}_k^i \leq \Delta_k^i \\
    & \|\delta \sigma^i \|_1 \leq \Delta_\sigma^i.
\end{align}
\end{subequations}

In the cost function, the term \( \sigma^i \) represents the time-scaling factor, which is minimized alongside the control inputs \( \bm{u}_k^i \).
The term \( \bm{\omega}_\nu \| \bm{\bar{\nu}}^i \|_1 \) penalizes the introduction of virtual control \( \bm{\nu}_k^i \), which is used to avoid infeasibility in the convexification process. 
The terms \( \bm{\omega}_{\bm{\Delta}}^i \| \bm{\bar{\Delta}}^i \|_2 \) and \( \bm{\omega}_{\bm{\Delta}_{\sigma}} \| \bm{\Delta}_\sigma \|_1 \) penalize deviations from the previous iterates for the state and time-scaling factors.
The boundary conditions ensure that the initial wet mass \( m_0^i = m_{\text{wet}} \), initial position \( \bm{r}_{{\mathcal{I}}, 0}^i = \bm{r}_{{\mathcal{I}}, i} \), initial velocity \( \bm{v}_{{\mathcal{I}}, 0}^i = \bm{v}_{{\mathcal{I}}, i} \), and initial angular velocity \( \bm{\omega}_{{\mathcal{B}}, 0}^i = \bm{\omega}_{{\mathcal{B}}, i} \) are satisfied.
At the final time step \( K \), the position \( \bm{r}_{{\mathcal{I}}, K}^i = 0 \), the velocity \( \bm{v}_{{\mathcal{I}}, K}^i = \bm{v}_{{\mathcal{I}}, f} \), the quaternion \( \bm{q}_{{\mathcal{B}} \leftarrow {\mathcal{I}}, K}^i = \bm{q}_{{\mathcal{B}} \leftarrow {\mathcal{I}}, f} \), and the angular velocity \( \bm{\omega}_{{\mathcal{B}}, K}^i = 0 \) are enforced, along with constraints that the thrust vector components along the body frame axes \( \bm{e}_2 \) and \( \bm{e}_3 \) are zero.
To enable multi-landing site selection, only the z component of the final state is constrained to be zero, while the x and y components are sampled.

The dynamics of the system are defined by the discrete-time state equation \( \bm{x}_{k+1}^i = \bar{\bm{A}}_k^i \bm{x}_k^i + \bar{\bm{B}}_k^i \bm{u}_k^i + \bar{\bm{C}}_k^i \bm{u}_{k+1}^i + \bar{\Sigma}_k^i \sigma^i + \bar{\bm{z}}_k^i + \nu_k^i \), where \( \bar{\bm{A}}_k^i \), \( \bar{\bm{B}}_k^i \), and \( \bar{\bm{C}}_k^i \) are the system matrices, \( \bar{\bm{z}}_k^i \) is a disturbance term, and \( \nu_k^i \) is the virtual control used to maintain feasibility. 
The state constraints ensure the mass is greater than or equal to the dry mass \( m_{\text{dry}} \).

Control constraints ensure that the thrust vector magnitude \( \bm{u}_k^i \) lies between \( T_{\text{min}} \) and \( T_{\text{max}} \), and the thrust direction is constrained by \( \cos \delta_{\text{max}} \| \bm{u}_k^i \|_2 \leq \bm{e}_1 \cdot \bm{u}_k^i \), where \( \delta_{\text{max}} \) defines the maximum allowable angle between the thrust and the body frame axis.
Trust regions are defined by bounds on the change in state \( \delta \bm{x}_k^i \) and control \( \delta \bm{u}_k^i \) to ensure the convex sub-problems remain bounded and feasible throughout the iteration process.
The constraints \( \delta \bm{x}_k^i \cdot \delta \bm{x}_k^i + \delta \bm{u}_k^i \cdot \delta \bm{u}_k^i \leq \Delta_k^i \) and \( \| \delta \sigma^i \|_1 \leq \Delta_\sigma^i \) enforce the trust regions around the previous iterate.

Outputs from solving Equations~\eqref{eq:cost_function}-\eqref{eq:trust_regions} are formulated as a 2D matrix to train the diffusion trajectory for our model:

\begin{equation}
    \mathbf{z} = \begin{bmatrix}
    \mathbf{r}_0 & \mathbf{r}_1 & \cdots & \mathbf{r}_N \\
    \mathbf{v}_0 & \mathbf{v}_1 & \cdots & \mathbf{v}_N \\
    \mathbf{q}_0 & \mathbf{q}_1 & \cdots & \mathbf{q}_N \\
    \mathbf{\omega}_0 & \mathbf{\omega}_1 & \cdots & \mathbf{\omega}_N \\
    m_0 & m_1 & \cdots & m_N \\
    \mathbf{T}_0 & \mathbf{T}_1 & \cdots & \mathbf{T}_N
    \end{bmatrix},
    \label{eq: trajectoryz}
\end{equation}

The states and control inputs ${\bm{z}}$ are stored over the horizon time $N$. Combining these matrices into a set of the required batch size, we have the diffusion state of the original trajectory $\bm{x}_0 = [{\bm{z}}^0_{0:N}, \; {\bm{z}}^1_{0:N}, \; \dots, {\bm{z}}^b_{0:N}]$, for batch size $b$.

\subsubsection{Flexible Constraint Enforcement using Compositional Diffusion}

To incorporate new constraints into the trajectory diffusion model defined in the previous section, composable diffusion can create high-energy regions where the constraints are violated in the resulting model.
We demonstrate compositional constraint enforcement by enforcing the glideslope constraint, as defined in Equation~\eqref{eq: glideslope}.

\begin{equation}
    \tan \gamma_{gs} \| H_\gamma \mathbf{r}_I(t) \|_2 \leq e_1 \cdot \mathbf{r}_I(t), \\
    \label{eq: glideslope}
\end{equation}

where $e_1$ is the $z$ component unit vector, $\gamma_{gs}$ is the glideslope angle, and $H_\gamma$ is the glideslope matrix.
Since composition via~\acp{ebm} requires differentiable energy functions, the violation of the glideslope constraint is defined as a smooth approximation of the ReLU function:
\begin{equation}
    \text{violation} = \log\left(1 + e^{\tan \gamma_{gs} \| H_\gamma \mathbf{r}_I(t) \|_2 - e_1 \cdot \mathbf{r}_I(t)}\right).
\end{equation}

An energy penalty associated with the glideslope constraint can then be computed as proportional to the square of the violation:
\begin{equation}
E_{\text{constraint}} = -\lambda_{\text{penalty}} \cdot \text{violation}^2,
\label{eq: constraint energy function}
\end{equation}

where $\lambda_{\text{penalty}}$ defines a scaling factor for energy function.
Equation~\eqref{eq: constraint energy function} imposes high energy values when constraint violation is low and low energy values when constraint violation is high, due to the inclusion of a negative sign.
The energy function is defined in this way to allow for negation composition, since negation composition was shown to improve upon product composition for powered descent guidance~\cite{briden2024compositional}.

Using negation composition for the generative trajectory diffusion model and the glideslope constraint energy, we obtain the composed energy function:
\[
E(\mathbf{x}_t, t) = \alpha_1 E_{\text{traj}}(\mathbf{x}_t, t) - \alpha_2 E_{\text{constraint}}(\mathbf{x}_t, t),
\]

where the weights are set as \( \alpha_1 = 1.3 \) and \( \alpha_2 = 0.3 \).
The approximate probability density function and score function for the composed model, as defined by the energy function, can then be computed by Equations~\eqref{eq: EBM}-\eqref{eq: EBM score}.

\subsubsection{State and Control Conditioning}

Alternative to composition, equality constraints on the state and control over the trajectory horizon can be directly enforced in the reverse diffusion process via inpainting, Equation~\eqref{eq: state constraint}.
This approach ensures constraint satisfaction but is limited to explicit equality constraints on the state or control values.
Inpainting is implemented by creating a mask on the constrained state and control indices and setting each state and action to satisfy the inpainting constraints, Equation~\eqref{eq: state constraint}, at each diffusion timestep.
This strategy is effective due to enforced local consistency during each denoising step~\cite{JannerDuEtAl2022}.

\subsubsection{Multi-Landing Site Selection}

Given a grid map of potential landing sites, where each cell is assigned a risk level $R_i \in [0, 1]$ for $i \in \{1 \cdots N\}$, where $N$ is the total number of cells and the total risk sums to 1, a probability distribution function describing the minimum-risk landing site constraint can be formulated, where the distribution of landing site risk is applied at the last timestep in the trajectory.
Since this risk map is independent of the 6~\ac{dof} minimum-fuel landing problem, it can be directly composed with the diffusion model trained on Equations~\eqref{eq:cost}-\eqref{eq:control_constraints1}.
This allows for the integration of landing site risk information into the trajectory generation process without additional retraining of the diffusion model.

\section{Results}

A multi-landing site 6~\ac{dof}  powered descent guidance diffusion model is trained and benchmarked in this work.
Samples drawn from the diffusion model are compared with samples directly from solving the same problem with~\ac{scvx}.
Flexible constraint enforcement is shown for enforcement of the glideslope constraint on the trajectory diffusion model during inference time.
Inpainting is then used to condition the composed diffusion model on a set of waypoint equality constraints.
Finally, a demonstration of compositional diffusion is shown for multi-landing site selection given risk maps.

\subsection{Six Degree-of-Freedom Powered Descent Diffusion Model}\label{sec: traj diffusion model}

The training dataset for the trajectory diffusion model is generated in Python using CVXpy and ECOS \cite{DiamondBoyd2016, AgrawalVerschuerenEtAl2018, DomahidiChuEtAl2013}.
The~\ac{scvx} algorithm is called iteratively using Equations~\eqref{eq:cost_function}-\eqref{eq:trust_regions}, using the ECOS solver to solve each~\ac{socp} subproblem~\cite{SzmukReynoldsEtAl2020}.
All variables and parameters use non-dimensional quantities, and the~\ac{scvx} parameters are equivalent to the parameters in Szmuk and A\c{c}{\i}kme\c{s}e, unless otherwise specified \cite{SzmukAcikmese2018}.
Table~\ref{tab:algorithm_settings} defines the powered descent guidance problem parameters, including $K=20$ time discretization nodes.
To generate a training dataset with a range of initial positions, velocities, orientations, angular velocities, initial mass, and control input ranges, uniform sampling was performed for each sample according to the distributions in Table~\ref{tab:sampled}. All variables use dimensionless units, as defined in \cite{SzmukAcikmese2018}.

\begin{table}
    \caption{\label{tab:algorithm_settings} Problem Parameters}
    \centering
    \begin{tabular}{ll}
        \hline
        \hline \\
        \textbf{Parameter} & \textbf{Value} \\
        \hline \\
        Gravity ($g$) & $- {\bm{e}_1}$ \\
        Flight Time Guess ($t_{f \text{guess}}$) & 3 \\
        Fuel Consumption Rate (${\alpha_m}$) & 0.01 \\
        Thrust to COM Vector (${r_{T, {\mathcal{B}}}}$) & -0.01 ${\bm{e}_1}$ \\
        Angular Moment of Inertia ($J_{\mathcal{B}}$) & 0.01 $I_{3\times 3}$ \\
        Number of discretization nodes ($K$) & 20 \\
        \hline
        \hline
    \end{tabular}
\end{table}

\begin{table}
    \caption{\label{tab:sampled} Sampled Dataset}
    \centering
    \begin{tabular}{ll}
        \hline
        \hline \\
        \textbf{Parameter} & \textbf{Sampled Trajectory Distribution} \\
        \hline \\
        Initial Z Position ($r_{z,0}$) & $\sim {\mathcal{U}}$ [1, 4] \\
        Initial X Position ($r_{x,0}$) & $\sim {\mathcal{U}}$ [-2, 2] \\
        Initial Y Position ($r_{y,0}$) & $\sim {\mathcal{U}}$ [-2, 2] \\
        Initial Z Velocity ($v_{z,0}$) & $\sim {\mathcal{U}}$ [-1, -0.5] \\
        Initial X Velocity ($v_{x,0}$) & $\sim {\mathcal{U}}$ [-0.5, -0.2] \\
        Initial Y Velocity ($v_{y,0}$) & $\sim {\mathcal{U}}$ [-0.5, -0.2] \\
        Initial Quaternion (${{\bm{q}}_{{\mathcal{B}},0}}$) & ${\sim}$ euler to q (0,  ${\mathcal{U}}$ [-30, 30], \\ & ${\mathcal{U}}$ [-30, 30]) \\
        Initial Z Angular Velocity ($\omega_{z,0}$) & 0 \\
        Initial X Angular Velocity ($\omega_{x,0}$) & $\sim {\mathcal{U}}$ [-20, 20] \\
        Initial Y Angular Velocity ($\omega_{y,0}$) & $\sim {\mathcal{U}}$ [-20, 20] \\
        Final X Position ($r_{x,N}$) & $\sim {\mathcal{U}}$ [-4, 4] \\
        Final Y Position ($r_{y,N}$) & $\sim {\mathcal{U}}$ [-4, 4] \\
        Wet Mass ($m_{\text{wet}}$) & $\sim {\mathcal{U}}$ [2, 5] \\
        Dry Mass ($m_{\text{dry}}$) & $\sim {\mathcal{U}}$ [0.1, 2] \\
        Max Gimbal Angle (${\delta_{\max}}$) & $\sim {\mathcal{U}}$ [10, 90] \\
        Max Thrust (${T_{\max}}$) & $\sim {\mathcal{U}}$ [3, 10] \\
        Min Thrust (${T_{\min}}$) & $\sim {\mathcal{U}}$ [0.01, 1] \\
        \hline
        \hline
    \end{tabular}
\end{table}

\subsubsection{Diffusion Model Architecture}

A timestep embedding is used to encode the current diffusion timestep $t$ into an input for the diffusion model~\cite{briden2024compositional}.
The neural network defining the parameterized energy function $E_{\bm{\theta}} ({\bm{x}},t)$ is a U-Net due to their exceptional performance in trajectory planning tasks, including the advantage of variable planning horizon lengths~\cite{JannerDuEtAl2022, kaiser2020model}.
Table~\ref{tab: Unet} shows the U-Net architecture for the parameterized energy function.
The convolutional blocks also include transpose, layer norm, and linear layers, which are not included in the table for brevity.

\begin{table}[h!]
\centering
\caption{U-Net Diffusion Model Architecture} \label{tab: Unet}
\begin{tabular}{ l l }
\hline
\hline \\
\textbf{Layer}                  & \textbf{Weight Shape}        \\
\hline \\
\textbf{Input Layer}            & \texttt{(None, H, W, C)}     \\
\textbf{conv2\_d}                    & \texttt{(3, 3, 32, 1)}       \\
\textbf{conv\_block 1/conv2\_d}      & \texttt{(3, 3, 1, 32)}       \\
\textbf{conv\_block 1/conv2\_d 1}    & \texttt{(3, 3, 32, 32)}      \\
\textbf{conv\_block 2/conv2\_d}      & \texttt{(3, 3, 16, 64)}      \\
\textbf{conv\_block 2/conv2\_d 1}    & \texttt{(3, 3, 64, 64)}      \\
\textbf{conv\_block 3/conv2\_d}      & \texttt{(3, 3, 32, 128)}     \\
\textbf{conv\_block 3/conv2\_d 1}    & \texttt{(3, 3, 128, 128)}    \\
\textbf{conv\_block 4/conv2\_d}      & \texttt{(3, 3, 64, 256)}     \\
\textbf{conv\_block 4/conv2\_d 1}    & \texttt{(3, 3, 256, 256)}    \\
\textbf{conv\_block 5/conv2\_d}      & \texttt{(3, 3, 512, 256)}    \\
\textbf{conv\_block 5/conv2\_d 1}    & \texttt{(3, 3, 256, 256)}    \\
\textbf{conv\_block 6/conv2\_d}      & \texttt{(3, 3, 256, 128)}    \\
\textbf{conv\_block 6/conv2\_d 1}    & \texttt{(3, 3, 128, 128)}    \\
\textbf{conv\_block 7/conv2\_d}      & \texttt{(3, 3, 128, 64)}     \\
\textbf{conv\_block 7/conv2\_d 1}    & \texttt{(3, 3, 64, 64)}      \\
\textbf{conv\_block 8/conv2\_d}      & \texttt{(3, 3, 64, 32)}      \\
\textbf{conv\_block 8/conv2\_d 1}    & \texttt{(3, 3, 32, 32)}      \\
\textbf{Output Layer}                & \texttt{(3, 3, 32, C)}       \\               \\ \hline \hline
\end{tabular}
\end{table}

The model includes a downsampling path and an upsampling path.
The downsampling path reduces the input's spatial resolution and increases the number of feature channels.
The upsampling path restores the original resolution while progressively combining high-level features from the path.
Downsampling includes convolutional layers with layer normalization and Swish activations, followed by a max-pooling layer.
Once the bottleneck is reached, transported convolutional layers are used for upsampling, and skip connections from the downsampling path are concatenated with the upsampled features at each level.
Cosine beta scheduling is used to ensure that \(\alpha_{\text{cumprod}}(t)\) starts close to 1 (at \(t = 0\)) and gradually decreases to 0 as \(t \to T\).
This work utilizes the same cosine beta scheduling process as~\cite{briden2024compositional} and we refer the reader to~\cite{nichol2021improved} for additional information on cosine beta scheduling.

\begin{figure}
    \centering
    \includegraphics[width=0.9\textwidth]{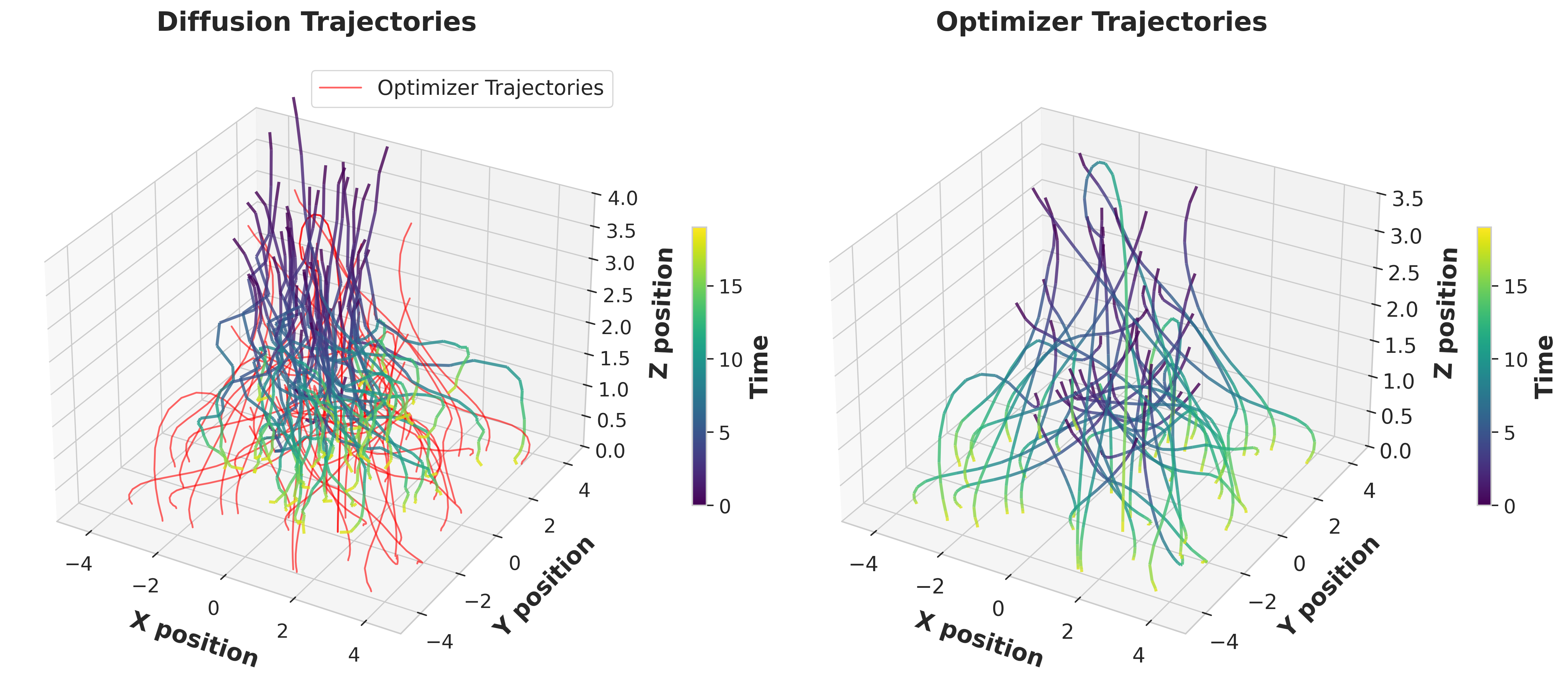}
    \caption{\revised{Diffusion model sampled trajectories and optimizer-generated trajectories for 6 DoF multi-landing site powered descent guidance.}}
    \label{fig:traj}
\end{figure}

\begin{figure}
    \centering
    \includegraphics[width=0.6\textwidth]{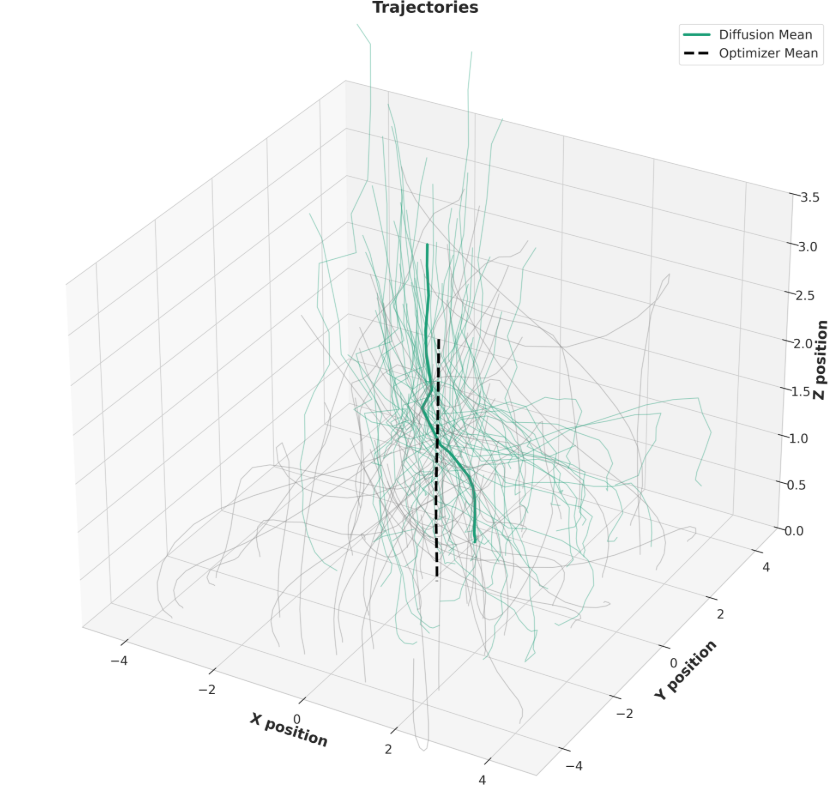}
    \caption{\revised{Comparison of the means and trajectory distributions for 1,000 diffusion model sampled trajectories and optimizer-generated trajectories for 6 DoF multi-landing site powered descent guidance. 50 out of the 1,000 sampled trajectories for each model are randomly selected and plotted with the means.}}
    \label{fig:trajmeanstd}
\end{figure}

\begin{figure}[htbp]
    \centering
    \includegraphics[width=0.6\textwidth]{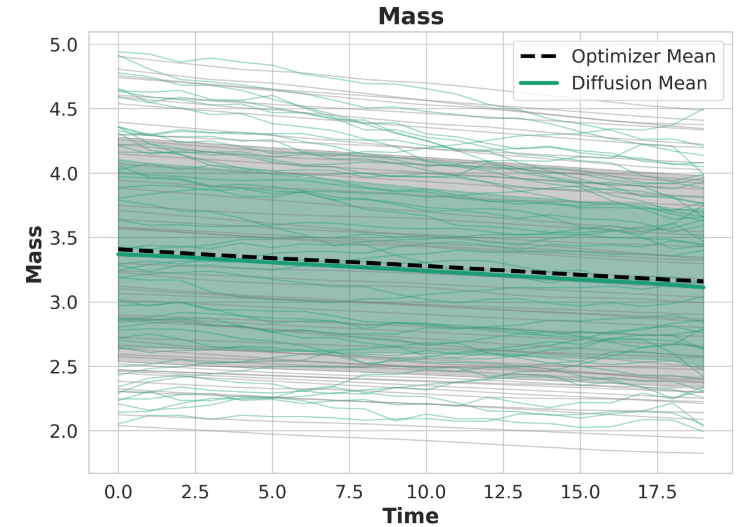}
    \caption{\revised{Mass Over Time. The mean values for the diffusion model's generated mass for 1,000 samples are shown in solid green, and the mean values for the optimizer's generated mass for 1,000 samples are shown in dotted black. The green and grey funnels show the 1-$\sigma$ bounds for the diffusion model mass and optimizer mass standard deviations. Fifty randomly selected trajectories from both generators are also shown in the same associated colors as the means and standard deviations.}}
    \label{fig:mass}
\end{figure}

\begin{figure}[htbp]
    \centering
    \begin{subfigure}[b]{0.5\textwidth}
        \centering
        \includegraphics[width=\textwidth]{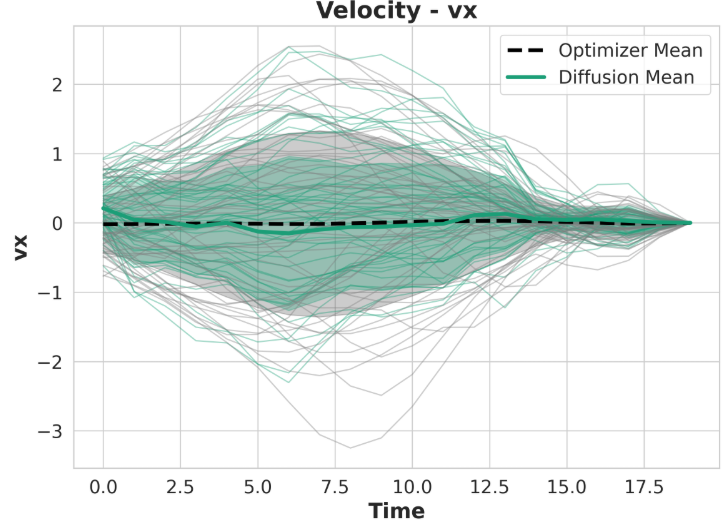}
        \caption{Velocity (X-component)}
        \label{fig:velocity_x}
    \end{subfigure}\hfill
    \begin{subfigure}[b]{0.5\textwidth}
        \centering
        \includegraphics[width=\textwidth]{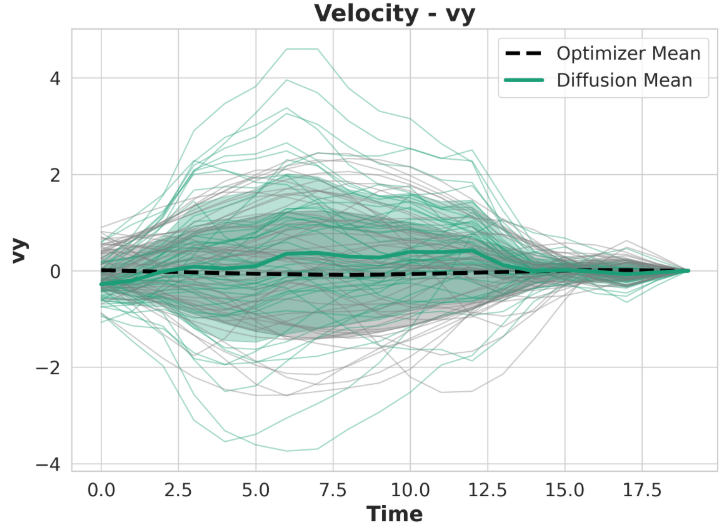}
        \caption{Velocity (Y-component)}
        \label{fig:velocity_y}
    \end{subfigure}
    \caption{\revised{Velocity Over Time (X and Y). The mean values for the diffusion model's generated velocity for 1,000 samples are shown in solid green, and the mean values for the optimizer's generated velocity for 1,000 samples are shown in dotted black. The green and grey funnels show the 1-$\sigma$ bounds for the diffusion model velocity and optimizer velocity standard deviations. Fifty randomly selected trajectories from both generators are also shown in the same associated colors as the means and standard deviations.}}
    \label{fig:velocity_xy}
\end{figure}

\begin{figure}[htbp]
    \centering
    \includegraphics[width=0.6\textwidth]{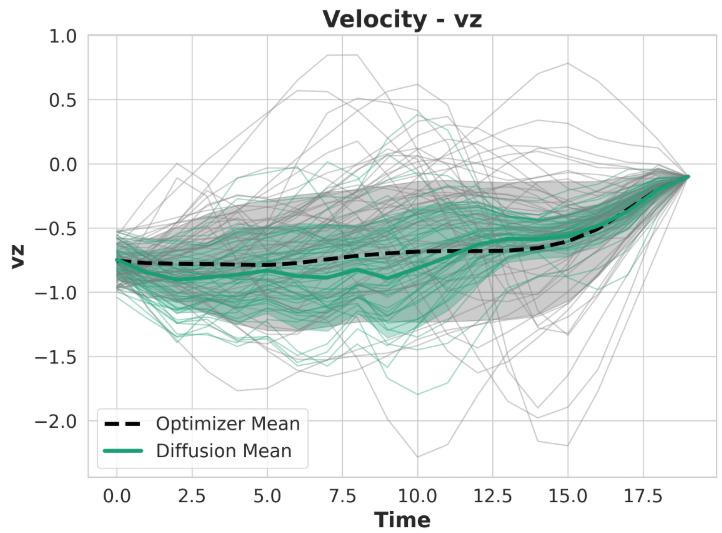}
    \caption{\revised{Velocity (Z-component) Over Time. The mean values for the diffusion model's generated velocity for 1,000 samples are shown in solid green, and the mean values for the optimizer's generated velocity for 1,000 samples are shown in dotted black. The green and grey funnels show the 1-$\sigma$ bounds for the diffusion model velocity and optimizer velocity standard deviations. Fifty randomly selected trajectories from both generators are also shown in the same associated colors as the means and standard deviations.}}
    \label{fig:velocity_z}
\end{figure}

\begin{figure}[htbp]
    \centering
    \begin{subfigure}[b]{0.5\textwidth}
        \includegraphics[width=\textwidth]{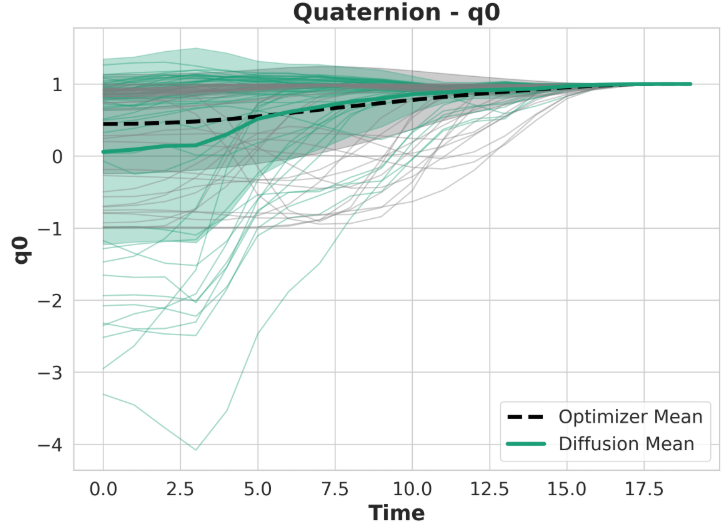}
        \caption{Quaternion q0}
        \label{fig:quaternion_q0}
    \end{subfigure}\hfill
    \begin{subfigure}[b]{0.5\textwidth}
        \includegraphics[width=\textwidth]{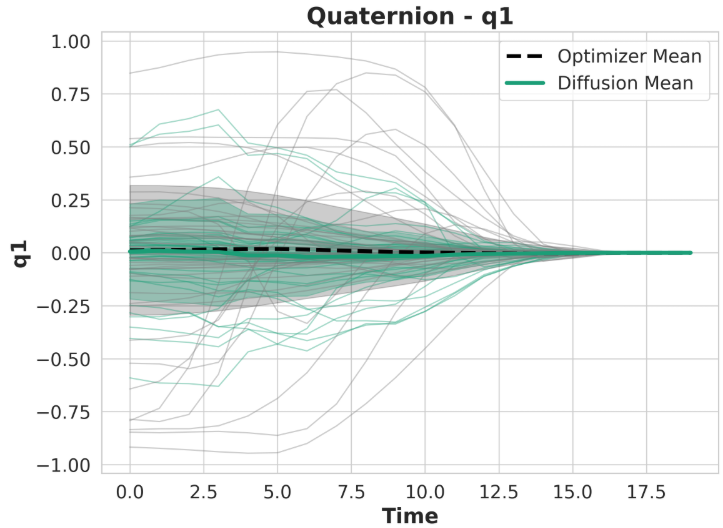}
        \caption{Quaternion q1}
        \label{fig:quaternion_q1}
    \end{subfigure}
    \caption{\revised{Quaternion q0 and q1 Over Time. The mean values for the diffusion model's generated quaternion component for 1,000 samples are shown in solid green, and the mean values for the optimizer's generated quaternion component for 1,000 samples are shown in dotted black. The green and grey funnels show the 1-$\sigma$ bounds for the diffusion model quaternion component and optimizer quaternion component standard deviations. Fifty randomly selected trajectories from both generators are also shown in the same associated colors as the means and standard deviations.}}
    \label{fig:quaternion_01}
\end{figure}

\begin{figure}[htbp]
    \centering
    \begin{subfigure}[b]{0.5\textwidth}
        \includegraphics[width=\textwidth]{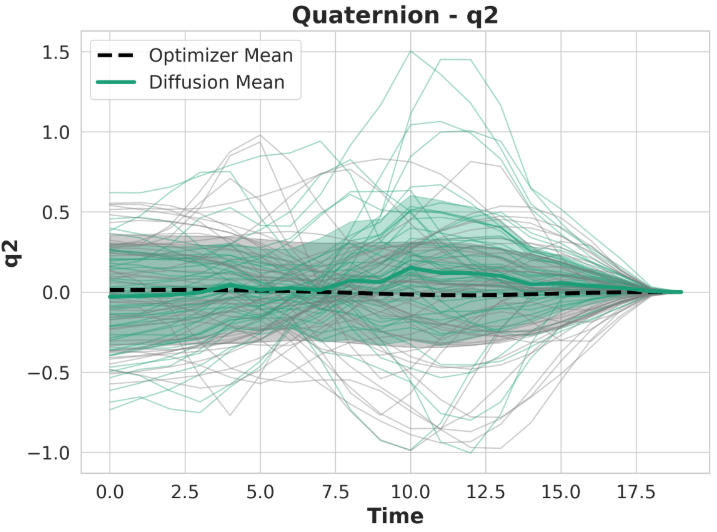}
        \caption{Quaternion q2}
        \label{fig:quaternion_q2}
    \end{subfigure}\hfill
    \begin{subfigure}[b]{0.5\textwidth}
        \includegraphics[width=\textwidth]{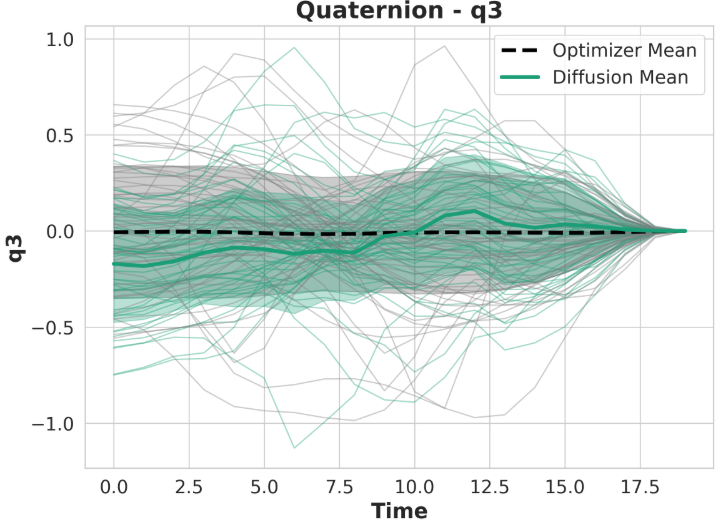}
        \caption{Quaternion q3}
        \label{fig:quaternion_q3}
    \end{subfigure}
    \caption{\revised{Quaternion q2 and q3 Over Time. The mean values for the diffusion model's generated quaternion component for 1,000 samples are shown in solid green, and the mean values for the optimizer's generated quaternion component for 1,000 samples are shown in dotted black. The green and grey funnels show the 1-$\sigma$ bounds for the diffusion model quaternion component and optimizer quaternion component standard deviations. Fifty randomly selected trajectories from both generators are also shown in the same associated colors as the means and standard deviations.}}
    \label{fig:quaternion_23}
\end{figure}

\begin{figure}[htbp]
    \centering
    \begin{subfigure}[b]{0.5\textwidth}
        \includegraphics[width=\textwidth]{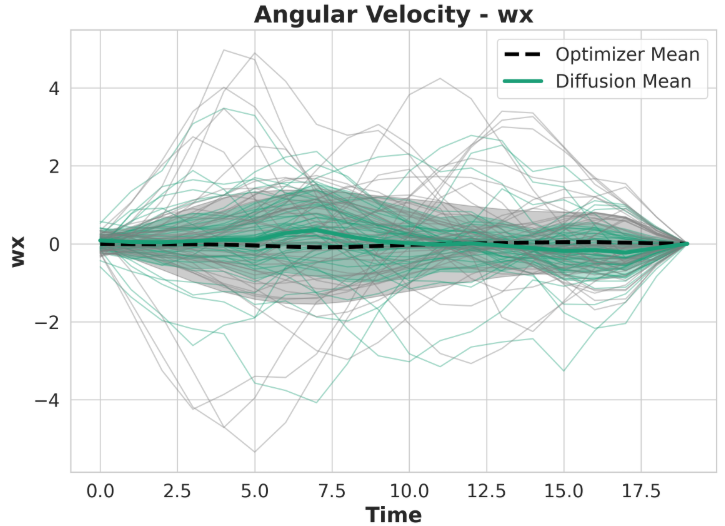}
        \caption{Angular Velocity (X)}
        \label{fig:angular_velocity_x}
    \end{subfigure}\hfill
    \begin{subfigure}[b]{0.5\textwidth}
        \includegraphics[width=\textwidth]{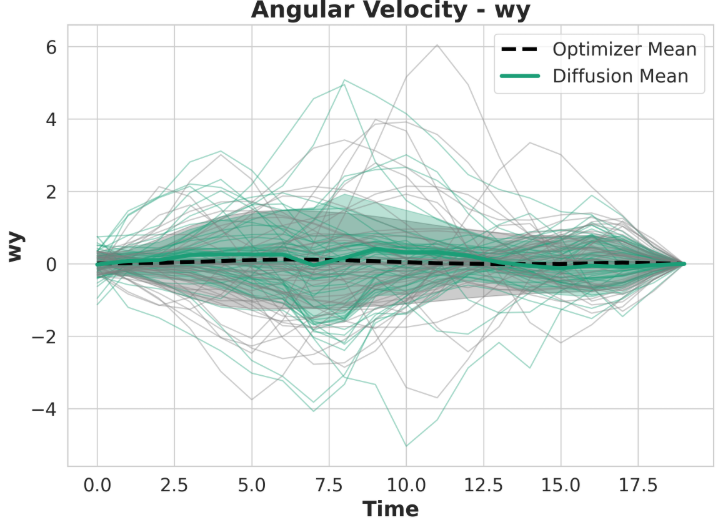}
        \caption{Angular Velocity (Y)}
        \label{fig:angular_velocity_y}
    \end{subfigure}
    \caption{\revised{Angular Velocity (X and Y) Over Time. The mean values for the diffusion model's generated angular velocity for 1,000 samples are shown in solid green, and the mean values for the optimizer's generated angular velocity for 1,000 samples are shown in dotted black. The green and grey funnels show the 1-$\sigma$ bounds for the diffusion model angular velocity and optimizer angular velocity standard deviations. Fifty randomly selected trajectories from both generators are also shown in the same associated colors as the means and standard deviations.}}
    \label{fig:angular_velocity_xy}
\end{figure}

\begin{figure}[htbp]
    \centering
    \includegraphics[width=0.6\textwidth]{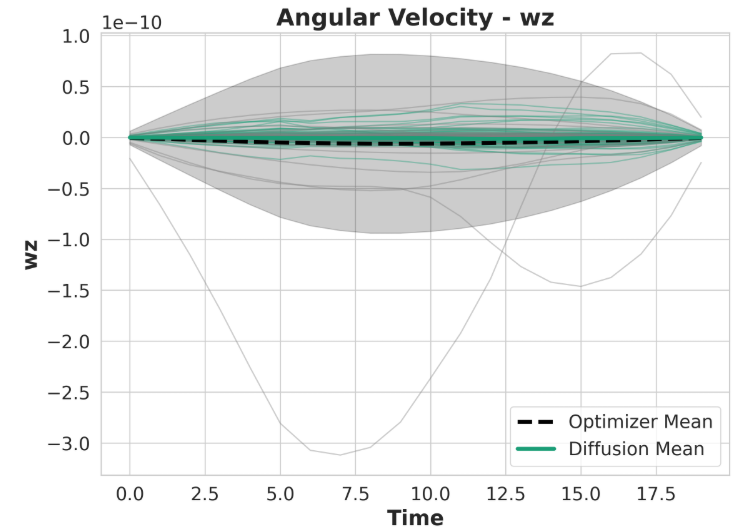}
    \caption{\revised{Angular Velocity (Z) Over Time. The mean values for the diffusion model's generated angular velocity for 1,000 samples are shown in solid green, and the mean values for the optimizer's generated angular velocity for 1,000 samples are shown in dotted black. The green and grey funnels show the 1-$\sigma$ bounds for the diffusion model angular velocity and optimizer angular velocity standard deviations. Fifty randomly selected trajectories from both generators are also shown in the same associated colors as the means and standard deviations.}}
    \label{fig:angular_velocity_z}
\end{figure}

\begin{figure}[htbp]
    \centering
    \begin{subfigure}[b]{0.5\textwidth}
        \includegraphics[width=\textwidth]{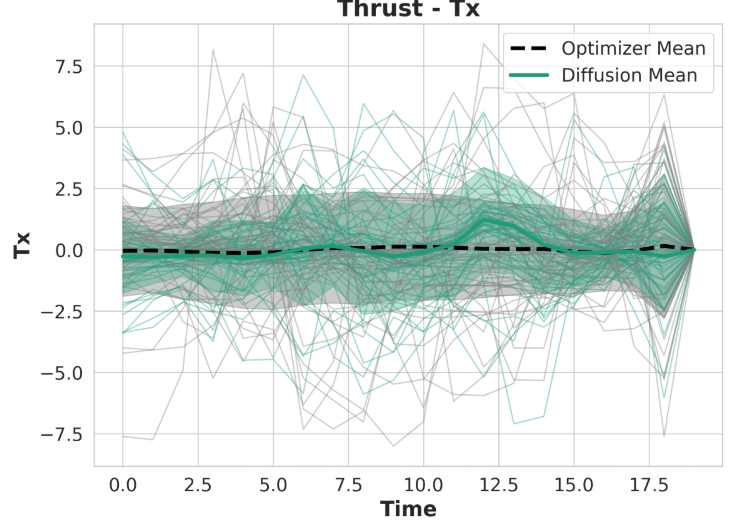}
        \caption{Thrust (X)}
        \label{fig:thrust_x}
    \end{subfigure}\hfill
    \begin{subfigure}[b]{0.5\textwidth}
        \includegraphics[width=\textwidth]{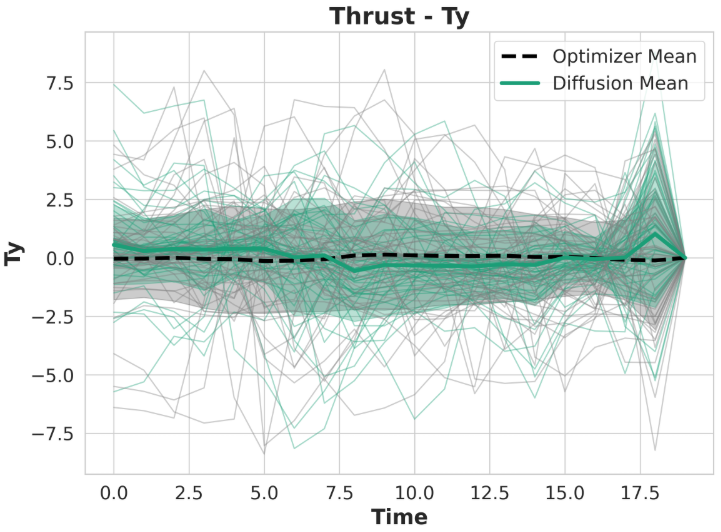}
        \caption{Thrust (Y)}
        \label{fig:thrust_y}
    \end{subfigure}
    \caption{\revised{Thrust (X and Y) Over Time. The mean values for the diffusion model's generated thrust for 1,000 samples are shown in solid green, and the mean values for the optimizer's generated thrust for 1,000 samples are shown in dotted black. The green and grey funnels show the 1-$\sigma$ bounds for the diffusion model thrust and optimizer thrust standard deviations. Fifty randomly selected trajectories from both generators are also shown in the same associated colors as the means and standard deviations.}}
    \label{fig:thrust_xy}
\end{figure}

\begin{figure}[htbp]
    \centering
    \includegraphics[width=0.6\textwidth]{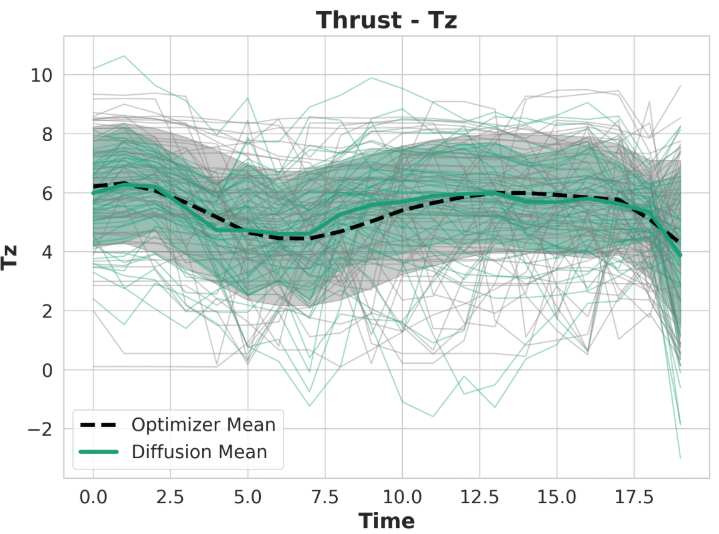}
    \caption{\revised{Thrust (Z) Over Time. The mean values for the diffusion model's generated thrust for 1,000 samples are shown in solid green, and the mean values for the optimizer's generated thrust for 1,000 samples are shown in dotted black. The green and grey funnels show the 1-$\sigma$ bounds for the diffusion model thrust and optimizer thrust standard deviations. Fifty randomly selected trajectories from both generators are also shown in the same associated colors as the means and standard deviations.}}
    \label{fig:thrust_z}
\end{figure}

For training and model outputs, values are scaled according to \texttt{sklearn.preprocessing.RobustScaler} fit on the training data~\cite{PedregosaVaroquauxEtAl2011}.
\texttt{RobustScaler} was used instead of \texttt{StandardScaler} since $x$ values outside of the range $[-1, 1]$ can significantly inhibit the learning process, and the samples obtained from the numerical optimizer often have outliers.
\revised{The training process used a batch size of 50 and was terminated after 3,320 training samples, when initialized using the TrajDiffuser single landing site diffusion model in~\cite{briden2024compositional}.
The complete model generates trajectories of batch size x 20 timesteps x 17 states.

Figures~\ref{fig:traj}-\ref{fig:thrust_z} compare diffusion model-generated and optimizer trajectories.}
Since diffusion trajectory samples are not conditioned on any initial or final states, they serve as a distributional comparison rather than an exact comparison.
The left plot in Figure~\ref{fig:traj} includes the diffusion model generated trajectories overlaid with the red optimizer trajectories.
The color scale shows the spacecraft's location over time, with 20 discretization nodes.
All trajectory points lie in the same distribution range as the sampled trajectories. 
Furthermore, a wide range of final positions are generated.

\revised{Providing further statistical information, Figure~\ref{fig:trajmeanstd} shows the mean and trajectory comparison for a larger range of trajectory samples, 1,000 samples from the diffusion model and 1,000 samples from the optimizer.
Generating 1,000 diffusion model samples required about 1.17 minutes while producing just 100 samples with the optimizer took about 5.67 minutes; the time estimate for computing 1,000 optimizer-generated trajectories is about one hour.
Figures~\ref{fig:mass}-\ref{fig:thrust_z} show the means and standard deviation bounds for all samples from the diffusion and optimizer-generated trajectories, showing 50 randomly selected trajectories for both diffusion and the optimizer.
Compared to the optimizer-generated trajectories, the diffusion model's trajectories are biased towards the negative x-direction by at most 0.7, biased towards the positive y-direction by less than 1.8, and biased slightly towards the positive z-direction, resulting in a 0.7 increase in starting position height.

Figure~\ref{fig:mass} shows the means and standard deviations for the optimizer and diffusion model-generated mass. 
When the two means are compared, the diffusion mean is only about 0.04 less than the optimizer mean.
This difference is also apparent in the optimizer's standard deviation, which encompasses a wider range of mass values.
While the diffusion model's generated mass is not always monotonically decreasing in Figure~\ref{fig:mass}, the range of masses between 2 and 5 are within the expected range of masses generated by the optimizer.
Further improvement could be achieved by increasing the number of training epochs or using a smoothing algorithm before using the mass as an initial guess to the optimizer.

The generated velocities in Figures~\ref{fig:velocity_xy}-\ref{fig:velocity_z} lie in the same -4 to 4 range of the optimizer's velocities.
While the diffusion model's mean generated velocities in the x-direction are primarily slightly less than the optimizer's, by less than 0.15, the diffusion velocities in the y-direction are greater by up to 0.5.
The larger y-direction velocities in the diffusion model are likely a result of the diffusion model trajectories guiding the spacecraft to final positions slightly outside the optimizer's trajectories in the y-direction.
This leads to a larger standard deviation in vy for the diffusion model, compared to the optimizer, and a smaller standard deviation in vx, compared to the optimizer.
The z-component of velocity, shown in Figure \ref{fig:velocity_z}, has a diffusion model mean at most 0.2 less than the optimizer's mean. 
Additionally, the standard deviation of the diffusion model's generated velocities grows smaller over time compared to the optimizer's standard deviation.
Still, the observed switching pattern between high and low z velocities occurs in both the generative model's and the optimizer's velocities.

Observing the generated quaternion model comparison in Figures~\ref{fig:quaternion_01}-\ref{fig:quaternion_23}, it is apparent that the diffusion quaternions exhibit the same policies as the optimizer-generated quaternions.
The q0 component starts with a lower diffusion-generated mean and a wider standard deviation than the optimizer. This pattern then switches five timesteps in, where the optimizer's standard deviation exceeds the diffusion model's.
Generally, this shows that the diffusion model encompasses a larger range of rotation starting angles.
The means and standard deviations for q1 and q2 almost exactly match, except for an increase in the diffusion model's mean and standard deviation after 7.5 timesteps.
Lastly, the q3 component starts with a lower mean and then switches to a higher mean after 10 timesteps.
Overall, the diffusion model means are within 0.2 of the optimizer means, and the diffusion model's standard deviations encompass a large region inside the optimizer's standard deviations.

Comparing angular velocities between the two models in Figures~\ref{fig:angular_velocity_xy}-\ref{fig:angular_velocity_z}, all means are almost identical, and the switching between x and y-directions are replicated in the diffusion model. The z-component of angular velocity, in Figure~\ref{fig:angular_velocity_z}, is zero due to the problem formulation.
Out of the x and y directions, the y direction has a slightly wider standard deviation for the diffusion model compared to the optimizer at around 7.5 timesteps. 
This further matches the larger y range in diffusion trajectory values.
The diffusion model's smaller x-direction angular velocity standard deviation results in less aggressive attitude changes.

The diffusion model thrust control input, in Figures~\ref{fig:thrust_xy}-\ref{fig:thrust_z} includes the same range of positive z-direction thrust values with x and y values centered around zero.
Alternations between the x-direction and y-direction thrusts are seen in both models as expected.
While the optimizer-generated thrusts have a mean centered around zero, the diffusion model's thrusts have a mean that reflects the switching behavior.
This mean is slightly biased towards a positive Tx at around 12 timesteps and a positive Ty at around 18 timesteps.
The z-direction thrust, shown in Figure~\ref{fig:thrust_z}, follows the same policy as the optimizer with only slightly lower endpoint thrusts.
To further expand the standard deviation of diffusion-generated thrusts, additional training with samples from the edge of the optimizer's 1-$\sigma$ region is recommended.}
For a full comparison of statistics for the dynamics constraint errors for a trained powered descent guidance diffusion model, we refer the reader to~\cite{briden2024compositional}.

\begin{figure*}[htbp]
    \centering
    \includegraphics[width=1\textwidth]{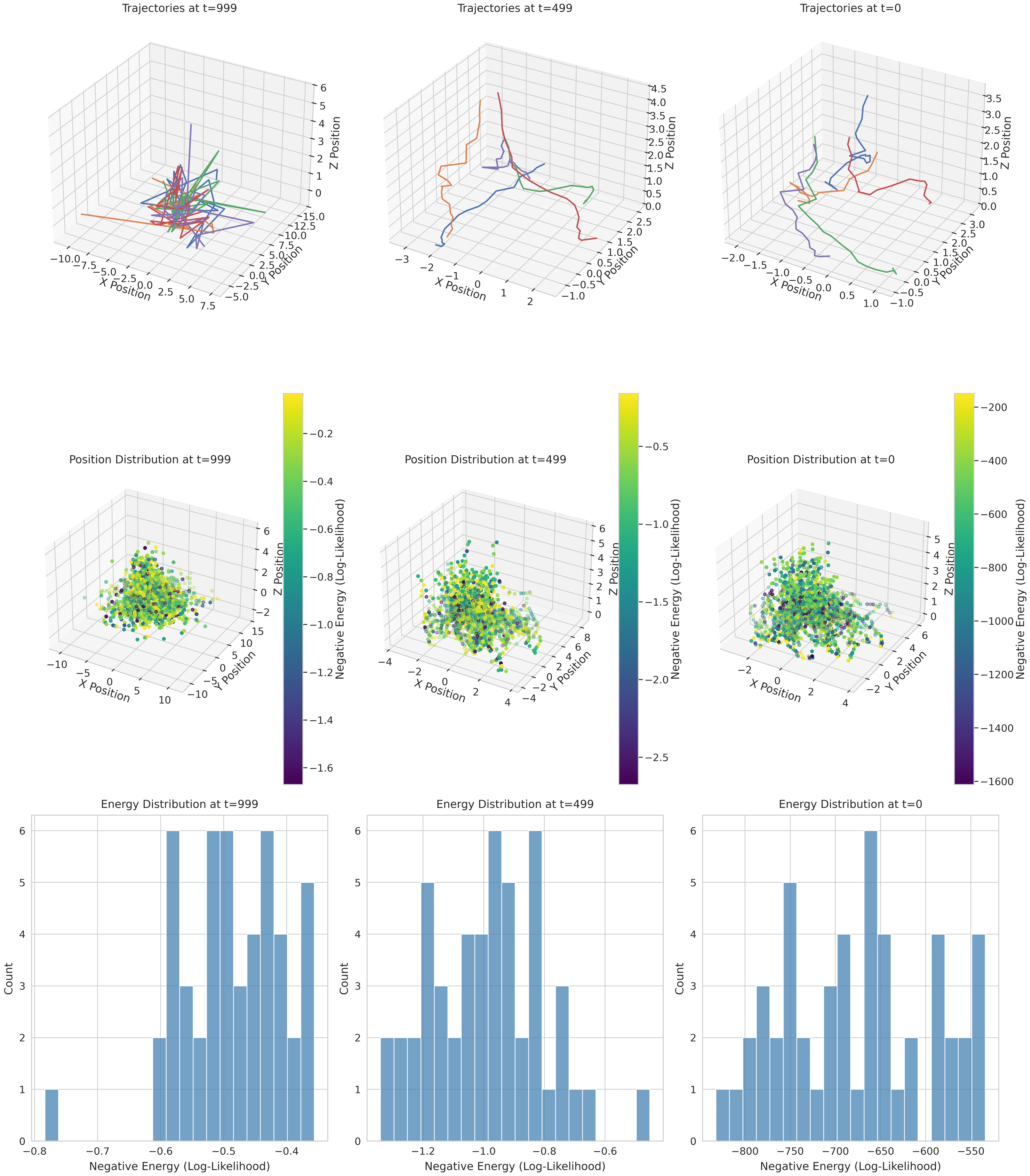}
    \caption{\revised{Trajectory and energy distributions for the trajectory diffusion model.}}
    \label{fig:traj_energy}
\end{figure*}

The~\ac{ebm} formulation for the compositional diffusion model, learning a parameterized energy function $E_{\bm{\theta}} (\bm{x}, t)$, provides insight into the negative log-likelihoods for the trajectory distribution. 
\revised{Given a set of reverse process trajectory samples $\bm{x}_t$ at the diffusion timestep $t$, the probability density function is proportional to $e^{-E_\theta (\bm{x}, t)}$, from Equation~\eqref{eq: EBM}. 
Practitioners are often interested in obtaining trajectory samples that are representative of the training data's distribution, as well as obtaining distributional information about that data. 
Using generative diffusion models, novel composed distributions can be obtained during inference time without requiring iterative runs of a representative number of newly formulated optimization problems. 
For the 6 DoF trajectory diffusion model trained in this work, we can plot and analyze the model's trained energy distribution, approximating the negative log-likelihoods. 
The negative energy, or approximate log-likelihood, at each trajectory discretization point and the distribution of negative energy values are shown in Figure~\ref{fig:traj_energy}.
The lefthand column of plots in Figure~\ref{fig:traj_energy} shows the standard normal distribution sample for a batch size of 5.
The middle row of plots shows the negative energy values for each point in 100 trajectory samples.
Low negative energy values are considered less likely data points, and high negative energy values are associated with a high likelihood of the generated trajectory containing that point.
Since the leftmost column negative energy position plot is from standard normal samples, no specific pattern occurs in the likelihoods.
The mean energy for each possible trajectory location is random, with only a 1.5-point range. 
The bottom row plots show the mean negative energy histogram, averaging over the state and time dimensions.
Analyzing the negative energy distribution for the standard normal sample, the negative energies are centered around -0.5.
The middle column plot shows the trajectories halfway through the reverse sampling process at timestep 499.
The negative energy now has a larger range of over 2, and low-energy clusters are forming.
The log-likelihoods are shifted to higher magnitudes with a similar distribution to the standard normal, moving further towards lower log-likelihoods.
Finally, the righthand column plot shows the sampled trajectories at the end of the reverse process, the desired trajectory output at timestep 999. 
Negative energy values are now much higher in magnitude and range by almost 1400 points.
The highest likelihood locations are primarily around landing sites, where there is less expected trajectory variation, and the lowest likelihood locations occur during the middle timesteps.}
The log-likelihood remains close to a Gaussian distribution but now has a thicker tail, sampling from a wide range of log-likelihoods.

\subsection{Flexible Constraint Enforcement using Compositional Diffusion}

\begin{figure*}[htbp]
    \centering
    \includegraphics[width=\textwidth]{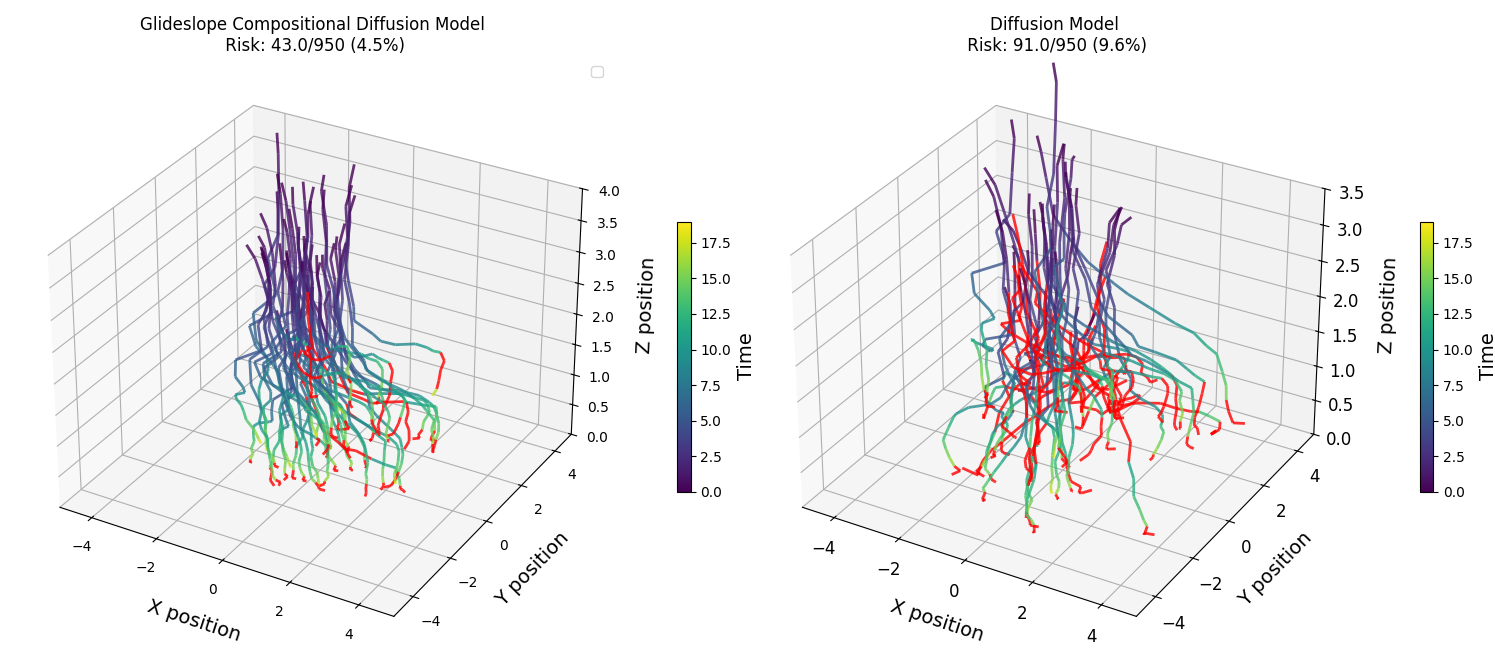}
    \caption{Glideslope constraint violation for the composed diffusion model vs. the trajectory diffusion model.}
    \label{fig:traj_glideslope}
\end{figure*}

Utilizing the constraint negation composition model described in Equation~\eqref{eq: constraint energy function}, the glideslope constraint is embedded in the sampling process during inference time.
No additional training is required to enforce this constraint.
Figure~\ref{fig:traj_glideslope} shows the number of segments that violate the glideslope constraint, with $\theta_\gamma = 30$, for the glideslope compositional diffusion model as compared to the trajectory diffusion model trained in Section~\ref{sec: traj diffusion model}.

Constraint violation is assessed for each trajectory segment, including 950 segments.
Out of these segments the composed model only violated the glideslope constraint in $4.5\%$ of the trajectories, often due to a non-smooth set of final points.
When the model is not composed, the diffusion model violates almost $10\%$ of the constrained segments.

\subsection{State and Control Conditioning}
\begin{figure*}[htbp]
    \centering
    \includegraphics[width=.5\textwidth]{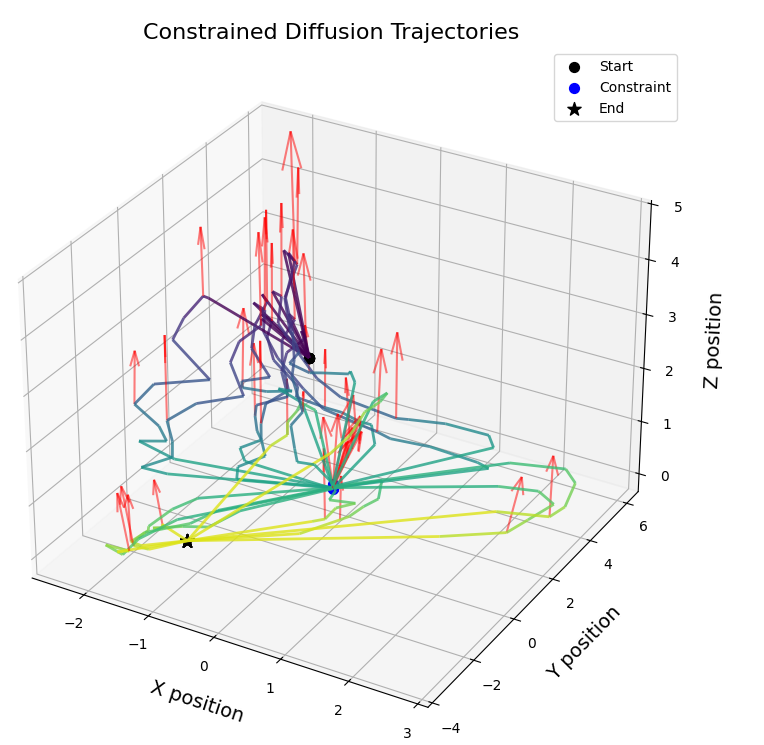}
    \caption{Constrained diffusion model generated trajectories, where the constraints include an initial state, intermediate state, and final state.}
    \label{fig:traj_constrained}
\end{figure*}

Figure~\ref{fig:traj_constrained} shows sampled trajectories using constrained waypoints in the reverse diffusion process via inpainting.
The red arrows indicate the thrust direction and magnitude at the discretization points.
The sampled diffusion trajectories exactly meet all equality constraints.
Additionally, provided trajectory options are multi-modal in that they do not follow only one direction, \ie{}, to the left or right of the next waypoint, but they provide approximately equal samples for both options.

\subsection{Multi-Landing Site Selection}

The process to create a risk map is defined by Algorithm~\ref{alg:risk_map} and the algorithm's parameters are detailed in Table~\ref{tab:parameters}.

\begin{algorithm}
\caption{Risk Map Generation with Elliptical Obstacles}
\label{alg:risk_map}
    \begin{algorithmic}[1]
        \Procedure{CreateRiskMap}{\texttt{obstacles}, $(x_{\min}, x_{\max})$, $(y_{\min}, y_{\max})$, \texttt{grid\_resolution}, $\sigma$}
            \State \textbf{Input:} List of elliptical obstacles \texttt{obstacles}, grid range $(x_{\min}, x_{\max})$ and $(y_{\min}, y_{\max})$, grid resolution \texttt{grid\_resolution}, Gaussian kernel standard deviation $\sigma$.
            \State \textbf{Output:} 2D grids $X, Y$ and normalized risk map \texttt{risk\_map}.
            \State Generate evenly spaced $x_{\text{vals}}$ and $y_{\text{vals}}$ using \texttt{grid\_resolution}.
            \State Create mesh grid $X, Y$ from $x_{\text{vals}}$ and $y_{\text{vals}}$.
            \State Initialize \texttt{risk\_map} as a zero matrix of shape $(X, Y)$.
            \For{\texttt{obs} in \texttt{obstacles}}
                \State Compute shifted coordinates relative to the obstacle center:
                $x_{\text{shift}} = X - h, \quad y_{\text{shift}} = Y - k$.
                
                \State Rotate grid points by the obstacle's angle \(\alpha\):
                
                $x_{\text{rot}} = x_{\text{shift}} \cos(\alpha) + y_{\text{shift}} \sin(\alpha)$,
                
                $y_{\text{rot}} = -x_{\text{shift}} \sin(\alpha) + y_{\text{shift}} \cos(\alpha)$.
                
                \State Compute the elliptical equation:
                $\text{ellipse\_eq} = \left(\frac{x_{\text{rot}}}{a}\right)^2 + \left(\frac{y_{\text{rot}}}{b}\right)^2$.
                
                \State Assign maximum risk ($1.0$) for grid points inside the ellipse ($\text{ellipse\_eq} \leq 1$).
                
                \State For points outside the ellipse ($\text{ellipse\_eq} > 1$), compute the distance:
                $\text{distance} = \sqrt{\text{ellipse\_eq}} - 1$.
                
                \State Calculate Gaussian risk:
                $\text{risk} = \exp\left(-\frac{\text{distance}^2}{2\sigma^2}\right)$.
                
                \State Add Gaussian risk to \texttt{risk\_map}.
            \EndFor
            \State Normalize \texttt{risk\_map} to $[0, 1]$
            
            using:
            $\texttt{risk\_map} \leftarrow \frac{\texttt{risk\_map}}{\max(\texttt{risk\_map})}$.
            \State \textbf{return} $X, Y, \texttt{risk\_map}$.
        \EndProcedure
    \end{algorithmic}
\end{algorithm}

\begin{table}[h!]
\centering
\caption{Input Parameters for Risk Map Generation}
\label{tab:parameters}
\begin{tabular}{@{}ll@{}}
\toprule
\textbf{Parameter} & \textbf{Description} \\ \midrule
\texttt{obstacles} & List of ellipses defined by \( a, b, h, k, \alpha \). \\
\texttt{x\_range} & Range of \( x \)-coordinates \((-4.0, 4.0)\). \\
\texttt{y\_range} & Range of \( y \)-coordinates \((-4.0, 4.0)\). \\
\texttt{grid\_resolution} & Spacing of the 2D grid (\( 0.1 \)). \\
\texttt{\(\sigma\)} & Standard deviation for Gaussian kernel (\( 0.5 \)). \\ \bottomrule
\end{tabular}
\end{table}

To ensure consistency across different obstacle configurations, the risk map is normalized:
\[
\text{normalized\_risk} = \frac{\text{risk}}{\max(\text{risk})}.
\]

The obstacle-avoiding diffusion model is defined through negation composition with a base trajectory diffusion model, as applied with the glideslope constraint compositional model:

\begin{equation}
p_{\text{avoid}}(\bm{x}) \propto \frac{p_{\text{traj}}(\bm{x})}{p_{\text{risk map}}(\bm{x})^{\alpha}},
\end{equation}

where $p_{\text{traj}}(\bm{x})$ is the base trajectory distribution and $p_{\text{risk map}}(\bm{x})$ is the obstacle risk map distribution.

The energy function for the composed model is:
\begin{equation}
E_{\text{composed}}(\bm{x}_t, t) = \alpha_1 E_{\text{traj}}(\bm{x}_t, t) - \alpha_2 E_{\text{obstacle}}(\bm{x}_t, t),
\end{equation}
where the obstacle energy term is defined using a risk map $R(x,y)$ and penalty function:
\begin{equation}
E_{\text{obstacle}}(\bm{x}_t, t) = \lambda \sum_{k \in K} w_k \cdot \max(0, -(R(x_k, y_k) - \tau)) \cdot s(t),
\end{equation}
where $\lambda$ is the penalty coefficient, $K$ are the key trajectory points (end points), $w_k$ are point-specific weights, $\tau$ is the risk threshold, $s(t)$ is the timestep scaling factor, and $R(x,y)$ is the risk map value at position $(x,y)$.

The risk map values are computed using bilinear interpolation:
\begin{equation}
R(x,y) = \sum_{i,j \in {0,1}} R_{x_i,y_j} \cdot (1-|x-x_i|) \cdot (1-|y-y_j|),
\end{equation}
where $(x_i,y_j)$ are the four nearest grid points.

The influence of obstacle avoidance varies across diffusion time using:
\begin{equation}
s(t) = \sin(\pi t/T),
\end{equation}
where $T$ is the total number of diffusion steps.

Key trajectory points are weighted differently to ensure a smooth gradient exists across the trajectory and to use local consistency to promote low risk landing site choices.
We found empirically that inducing a higher weight on the intermediate points improved constraint satisfaction compared to only enforcing the~\ac{ebm} on the final position.
Point weights used for this analysis are as follows:
\begin{equation}
w_k = \begin{cases}
1.0 & \text{for start/end points} \\
2.0 & \text{for middle points}.
\end{cases}
\end{equation}

The score function for sampling is the gradient of the energy can then be computed by taking the gradient of the log-likelihood,
\begin{equation}
\nabla_x \log p_{\text{avoid}}(x_t|t) = -\nabla_x E_{\text{composed}}(x_t, t).
\end{equation}

\begin{figure*}[t!]
    \centering
    \includegraphics[width=\textwidth]{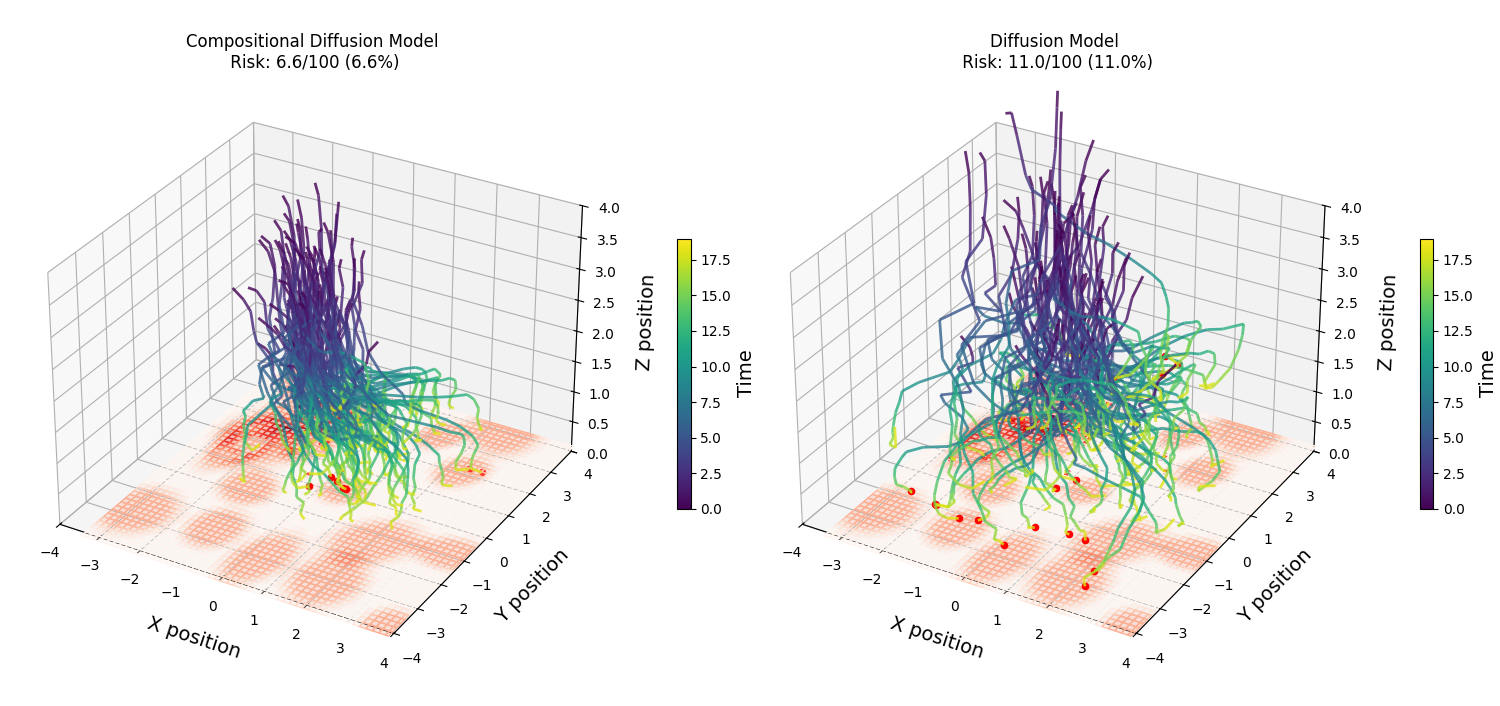}
    \caption{Landing risk for the composed diffusion model vs. the trajectory diffusion model.}
    \label{fig:risk}
\end{figure*}

\begin{figure*}[h!]
    \centering
    \includegraphics[width=1\textwidth]{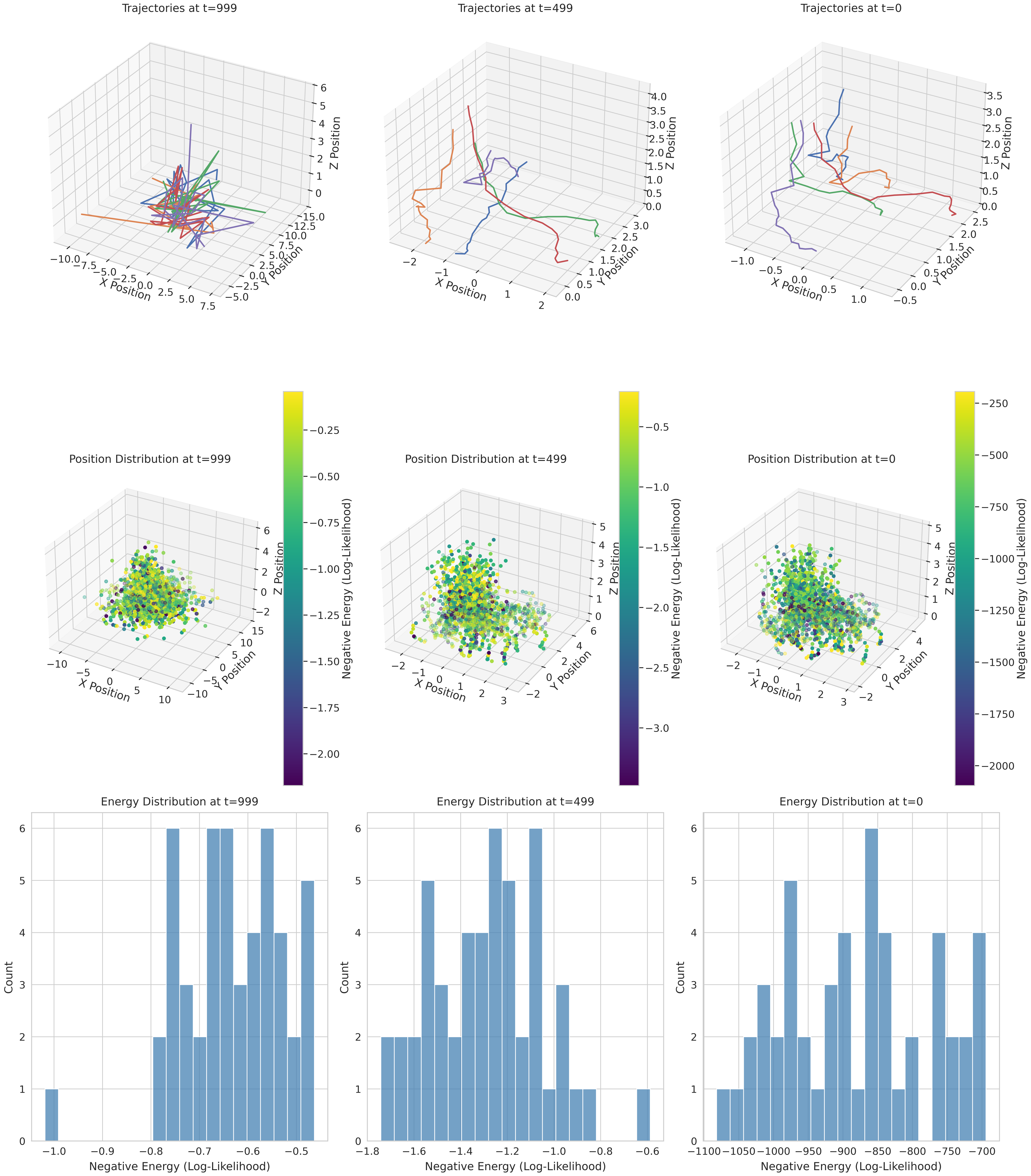}
    \caption{\revised{Trajectory and energy distributions for the composed trajectory diffusion model.}}
    \label{fig:traj_energy_composed}
\end{figure*}

Figure~\ref{fig:risk} shows the landing risk for a sampled risk map for the composed diffusion model versus the trajectory diffusion model trained in Section~\ref{sec: traj diffusion model}.
Trajectories that reach landing sites with $>10\%$ risk are marked with red dots.
Compared to the $11\%$ constraint violation in the trajectory diffusion model, the composed model generates trajectories with less than $7\%$ landing in regions with over $10\%$ risk.
If~\ac{mcmc} is employed for some of the sample steps we expect further improvements, with the reverse diffusion process representing an upper bound on the error rate associated with constrained trajectory sampling during inference time.

\revised{Figure~\ref{fig:traj_energy_composed} shows the trajectories and negative energy, or approximate log-likelihoods, over the reverse sampling process, which can be directly compared to the distributions shown in the trajectory-only diffusion model, Figure~\ref{fig:traj_energy}.
While the standard normal samples, in the left-hand column, and the intermediate samples, in the middle column, are very similar in appearance and negative energy to Figure~\ref{fig:traj_energy}, the final trajectory samples are substantially different. 
The much higher magnitude likelihoods still appear Gaussian. 
While the trajectories look similar in dynamics, the landing sites vary fairly widely from the trajectory diffusion model.
Overall, we can observe a shift in the distribution, occurring primarily in the final distribution when energy-based diffusion models are composed.}

\section{Discussion}

\revised{
By training a multi-landing site trajectory diffusion model, compositional constraint and cost function formulations enable the practitioner to construct a toolbox of building block models to achieve efficient initial guess generation for a generalized set of trajectory generation problems. 
The trained 6 DoF diffusion model was compared statistically to optimizer-generated trajectory samples.
Negation-based composition of the trajectory diffusion model with a differentiable glideslope constraint was achieved using reverse diffusion.
We found that negation composition with reverse diffusion improved constraint satisfaction over product composition, as shown experimentally in~\cite{briden2024compositional}.
While not shown theoretically, it is likely that reverse diffusion product composition, which adds the score functions, is more sensitive to differences in the magnitude of the composed model's score functions.
As such, if one score function is significantly larger than the other, the composition will not affect the generated trajectories.
Since negation using reverse diffusion approximately removes the areas of low energy for regions that violate the constraints, disparities in score function magnitudes may have less impact on the constraint satisfaction for generated trajectories.
Using negation composition during inference time resulted in a less than 5\% constraint violation. 
When constraints can be directly conditioned on state or control, inpainting can generate dynamically feasible trajectories that satisfy the specified equality constraints. 
This work shows this using waypoint constraints.
Finally, a negation compositional diffusion model was constructed for the probabilistic multi-landing site problem, reducing constraint violations by 40\%. 
Since energy-based diffusion models enable the approximate representation of the negative log-likelihood for the diffusion model, these distributions were compared for the trajectory generative diffusion model and the composed multi-landing site diffusion model.
The results showed a shift in the associated log-likelihoods for potential landing site locations that match the new final trajectories.}

\section{Conclusion}

This work has demonstrated multi-modal trajectory generation with flexible constraint conditioning using compositional diffusion models.
A multi-landing site generative diffusion model was trained for 6~\ac{dof} powered descent guidance. 
Constraint-abiding trajectories were generated during inference time using negation composition and inpainting without additional training.
Diffusion-generative models are efficient and adaptable compared to alternative initial guess-generating techniques, making them an enabling technology for autonomous space exploration.



\section*{Acknowledgments}
This work was supported in part by a NASA Space Technology Graduate Research Opportunity 80NSSC21K1301.
This research was carried out in part at the Jet Propulsion Laboratory, California Institute of Technology, under a contract with the National Aeronautics and Space Administration and funded through the internal Research and Technology Development program.

\bibliography{main}

\end{document}

%% file: definitions.tex
\newacronym[longplural={quadratic programs}]{qp}{QP}{quadratic program}
\newacronym[longplural={second-order cone program}]{socp}{SOCP}{second-order cone program}
\newacronym[longplural={interior point methods}]{ipm}{IPM}{interior point method}
\newacronym[longplural={sequential quadratic programs}]{sqp}{SQP}{sequential quadratic program}
\newacronym[longplural={sequential convex programs}]{scp}{SCP}{sequential convex program}
\newacronym[longplural={nonlinear programs}]{nlp}{NLP}{nonlinear program}

\newacronym{scvx}{SCvx}{Successive Convexification}
\newacronym{lcvx}{LCvx}{Lossless Convexification}
\newacronym{tscvx}{T-SCvx}{Transformer-based Successive Convexification}
\newacronym{upg}{UPG}{Universal Powered Guidance}
\newacronym{pdg}{PDG}{Powered Descent Guidance}
\newacronym{tpdg}{T-PDG}{Transformer-based Powered Descent Guidance}

\newacronym[longplural={degrees-of-freedom}]{dof}{DoF}{degree-of-freedom}
\newacronym{pdf}{PDF}{probability density function}
\newacronym[longplural={energy-based models}]{ebm}{EBM}{energy-based model}
\newacronym{mcmc}{MCMC}{Markov Chain Monte Carlo}
\newacronym{ula}{ULA}{unadjusted Langevin Dynamics}

\newacronym[longplural={Denoising Diffusion Probabilistic Models}]{ddpm}{DDPM}{Denoising Diffusion Probabilistic Model}
\newacronym[longplural={Deep Neural Networks}]{dnn}{DNN}{Deep Neural Network}

\newacronym{ido}{IDO}{Iterative Diffusion Optimization}
\newacronym{socm}{SOCM}{stochastic optimal control matching}

\newcommand{\eg}{{e.g.}}
\newcommand{\ie}{{i.e.}}